\documentclass[letterpaper]{article} 
\usepackage[preprint]{aaai2027}  
\usepackage[hyphens]{url}  
\usepackage{graphicx} 
\usepackage{natbib}  
\usepackage{caption} 
\usepackage{algorithm}
\usepackage{algorithmic}

\usepackage{newfloat}
\usepackage{amsmath}
\usepackage{amssymb}
\usepackage{listings}
\DeclareCaptionStyle{ruled}{labelfont=normalfont,labelsep=colon,strut=off} 
\floatstyle{ruled}
\newfloat{listing}{tb}{lst}{}
\floatname{listing}{Listing}

\usepackage{booktabs}
\usepackage[table]{xcolor}
\title{TAPD: TTS-Guided Acoustic Policy Distillation for Paralinguistic Generation in Speech Language Models}
\author{
    Tianyu Peng\textsuperscript{\rm 1,\rm 2},
    Yingzhuo Deng\textsuperscript{\rm 1,\rm 2},
    Jiajun Zhang\textsuperscript{\rm 1,\rm 2}\corresponding
}
\affiliations{
    \textsuperscript{\rm 1}Institute of Automation, Chinese Academy of Sciences\\
    \textsuperscript{\rm 2}School of Artificial Intelligence, University of Chinese Academy of Sciences
}

\usepackage{tabularx}
\usepackage{array}
\usepackage{fvextra}
\usepackage{CJKutf8}
\pdfmapfile{+gbsnu.map}
\DefineVerbatimEnvironment{PromptBlock}{Verbatim}{
    fontsize=\scriptsize,
    breaklines=true,
    breakanywhere=true,
    xleftmargin=1em,
    xrightmargin=1em
}

\begin{document}
\begin{CJK*}{UTF8}{gbsn}

\pagestyle{plain}
\maketitle

\begin{abstract}
TTS-synthesized speech has become a common source of supervision for fine-tuning speech language models.
However, this kind of TTS-synthetic SFT treats TTS models only as offline teachers and provides no feedback on the student's online generation, leaving the student's on-policy acoustic errors such as flattened emotion, unstable speaking rate, and inappropriate loudness uncorrected.
We propose \textbf{TAPD}, a TTS-guided acoustic policy distillation framework that uses a frozen controllable TTS model as a lightweight acoustic teacher for speech LLM (student) post-training.
After each student rollout, TAPD evaluates the cumulative speech prefix at each audio-chunk boundary and assigns the tail-region teacher-forcing NLL associated with the newly appended chunk as localized feedback.
This feedback is mapped to the student's audio-token policy without requiring shared tokenization, output space, or teacher logits, enabling heterogeneous acoustic distillation.
Experiments on multiple paralinguistic dimensions with two TTS teachers show that TAPD consistently achieves the best student-side average instruction-following rate.
It outperforms matched TTS-synthetic SFT by 5.53 and 5.06 percentage points under the two teacher settings, respectively.
Score validation and ablation studies further confirm the effectiveness of TTS-derived acoustic feedback, chunk-level credit assignment, and selective distillation.
\end{abstract}

\section{Introduction}
\label{sec:introduction}

End-to-end speech language models
\cite{DBLP:conf/emnlp/ZhangLZZWZQ23,DBLP:journals/corr/abs-2412-02612,DBLP:journals/corr/abs-2410-00037,DBLP:journals/corr/abs-2507-16632}
have recently advanced spoken dialogue generation by directly modeling speech inputs and outputs without explicit ASR--LLM--TTS cascades.
However, semantic correctness is only one aspect of spoken communication:
a capable assistant should also control emotion, speaking rate, loudness, and composite speaking styles.
Current speech LLMs may still produce flattened emotion, unintended speaking rates, inappropriate loudness, or incomplete realizations of multiple style constraints
\cite{DBLP:journals/corr/abs-2510-22588}.
A practical solution is to synthesize controllable speech with modern TTS systems
\cite{DBLP:journals/corr/abs-2412-10117,DBLP:journals/corr/abs-2505-17589}
for speech LLM fine-tuning
\cite{DBLP:journals/corr/abs-2412-02612,DBLP:conf/emnlp/WangPYBWLJWWZ025},
as expressive conversational speech is expensive to collect and annotate
\cite{DBLP:conf/mm/Jin0W0ZZQ024}.
Nevertheless, TTS-synthetic SFT remains an offline imitation objective:
the student learns from fixed reference speech but receives no feedback on its own generations, allowing paralinguistic attributes to drift over extended autoregressive trajectories.

\begin{figure}[t]
    \centering
    \includegraphics[width=0.9\linewidth]{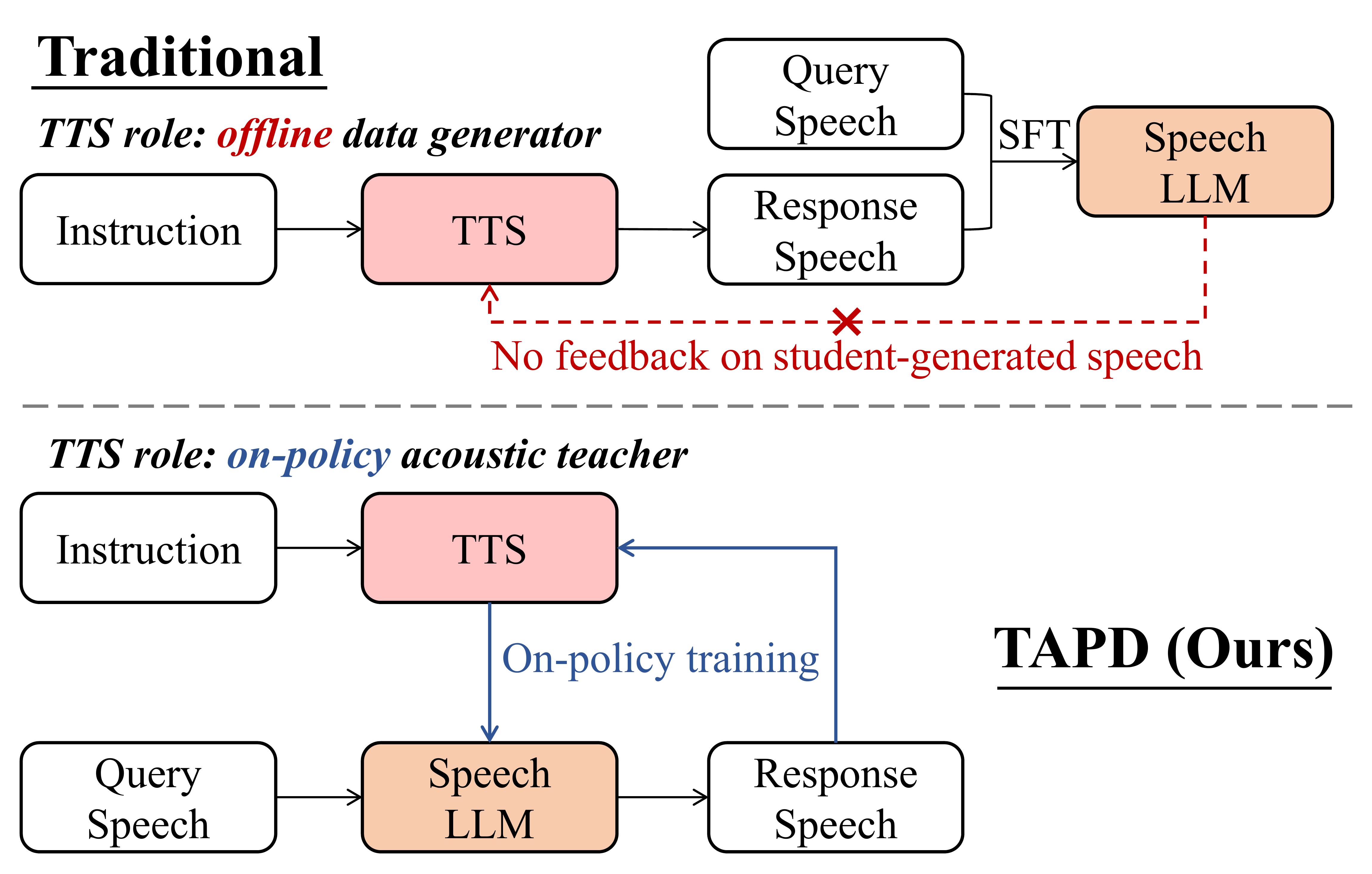}
    \caption{
    Overview and motivation of TAPD.
    }
    \label{fig:first}
\end{figure}

Likelihood-based scoring with pretrained speech generative models has been explored for evaluating generated speech \cite{DBLP:conf/icassp/MaitiPSW23}. 
In this work, we argue that controllable TTS models can play a stronger role.
Beyond synthesizing offline training data, they can serve as lightweight acoustic teachers for speech LLM post-training.
As illustrated in Figure~\ref{fig:first}, TAPD repurposes a frozen controllable TTS model from an offline data generator into an on-policy acoustic teacher for student-generated speech.
Under teacher forcing, the likelihood assigned by an instruction-conditioned TTS model can serve as a conditional acoustic compatibility signal, indicating how well the generation of speech LLMs agrees with the intended text, reference voice, and speaking style rather than merely measuring generic speech quality.
Moreover, because this feedback does not expose teacher-side audio-token logits, TAPD requires no shared tokenization, output space, or generation architecture, enabling heterogeneous acoustic distillation between the TTS teacher and speech LLM student.

Based on this insight, we propose \textbf{TAPD}, a \textbf{TTS-guided acoustic policy distillation} framework for on-policy speech LLM post-training.
The speech LLM student first generates a spoken response using its own policy.
After the rollout, TAPD evaluates the cumulative speech prefix at each audio-chunk boundary using a frozen controllable TTS teacher conditioned on the complete rollout text, reference prompt speech, and a cached style instruction. 
For each prefix, only the tail-region likelihood associated with the newly appended chunk is retained as localized acoustic feedback. 
This post-rollout cumulative-prefix scoring preserves the preceding acoustic context while providing localized supervision along the student’s own generation trajectory.

The resulting TTS signal provides chunk-level feedback rather than exact supervision for individual audio tokens.
Densely applying it to the full trajectory may therefore disrupt reliable token decisions.
TAPD instead applies acoustic policy distillation only to high-entropy audio-token positions, focusing the coarse-grained feedback on decisions where the student is uncertain.

We evaluate TAPD on the emotion, speaking-rate, loudness, and composite-style subsets of UltraVoice \cite{DBLP:journals/corr/abs-2510-22588} using two controllable TTS teachers.
Under both teacher settings, TAPD achieves the best student-side average IFR, improving over matched controllable-TTS SFT by 5.53 and 5.06 percentage points, respectively.
The largest gains are observed on composite styles, suggesting that localized on-policy acoustic feedback is particularly valuable when multiple paralinguistic constraints must be maintained throughout autoregressive generation.
Score validation and ablation studies further verify the effectiveness of TTS-derived acoustic feedback, chunk-level credit assignment, and entropy-based selective distillation.

Our contributions are summarized as follows:
\begin{itemize}
    \item We introduce controllable TTS models as lightweight acoustic teachers that score speech generated under the student's own policy, extending their role beyond offline data synthesis.

    \item We propose \textbf{TAPD}, combining cumulative prefix scoring, chunk-level tail attribution, and entropy-based selective optimization to perform heterogeneous acoustic policy distillation without shared tokenization or teacher logits.

    \item Experiments with two controllable TTS teachers demonstrate consistent improvements in paralinguistic controllability over strong offline and on-policy baselines.
\end{itemize}

\section{Related Work}

\subsection{Speech LLMs and TTS-Based Supervision}

End-to-end speech language models directly generate discrete audio tokens,
enabling spoken interaction without an explicit ASR--LLM--TTS
cascade \cite{DBLP:conf/emnlp/ZhangLZZWZQ23,DBLP:journals/corr/abs-2412-02612,DBLP:journals/corr/abs-2507-16632}.
Recent systems support instruction-conditioned control over
emotion, speaking rate, loudness, and other vocal attributes
\cite{DBLP:journals/corr/abs-2505-17589,DBLP:journals/corr/abs-2510-22588}.
Because expressive conversational speech is expensive to collect,
controllable TTS models are commonly used to synthesize
instruction-following speech for supervised fine-tuning.
Such approaches, however, treat TTS models as offline data
generators and train the student to imitate fixed synthetic
references, without evaluating acoustic errors that arise along its
own autoregressive generation trajectory.

A complementary line of work uses learned quality, style, or preference models to assess generated speech
\cite{DBLP:conf/icassp/MaitiPSW23,yang2026parass,DBLP:journals/corr/abs-2602-13891}.
Despite different architectures, these approaches generally collapse assessment into an utterance-, turn-, or episode-level signal, providing limited temporal credit assignment for localized acoustic errors.
In contrast, TAPD uses the conditional teacher-forcing likelihood of a frozen controllable TTS model as localized acoustic feedback on student-generated speech, without training an additional reward model.

\subsection{Cross-Tokenizer Distillation}

Standard token-level knowledge distillation \cite{DBLP:journals/corr/HintonVD15} generally assumes aligned
teacher and student vocabularies, making direct distribution matching
difficult when their tokenizations differ.
Sequence-level distillation avoids token-wise alignment by training
the student on teacher-generated responses, but mainly provides
offline response-level supervision \citep{DBLP:conf/emnlp/KimR16}.
Approximate Likelihood Matching (ALM) instead identifies teacher and
student token chunks corresponding to the same underlying text spans
and matches their sequence likelihoods, enabling distillation across
substantially different tokenizers \citep{DBLP:conf/nips/MinixhoferVP25}.
Other cross-tokenizer approaches align mismatched predictions through
vocabulary mappings, shared text spans, or intermediate
representations \citep{DBLP:journals/tmlr/BoizardHHC25,DBLP:conf/acl/ChenLM0XZ25,DBLP:conf/acl/DaoNCNVDL26}.
On-policy distillation instead evaluates teacher feedback on
student-generated trajectories, reducing the mismatch between fixed
distillation data and the student's inference-time distribution
\cite{DBLP:conf/iclr/AgarwalVZSGGB24}.
Recent work further extends on-policy logit distillation across model
families through explicit cross-tokenizer alignment
\cite{DBLP:journals/corr/abs-2606-09456}.

TAPD targets a more heterogeneous acoustic setting, where the TTS teacher and speech LLM student differ in tokenization, output space, token rate, and architecture. Using decoded waveforms as a shared interface, it scores cumulative speech prefixes and attributes newly introduced teacher-side likelihood regions to the corresponding student audio chunks, enabling localized distillation without explicit token-level alignment.

\section{Method}
\label{sec:method}

\begin{figure*}[t]
\centering
\includegraphics[width=0.9\textwidth]{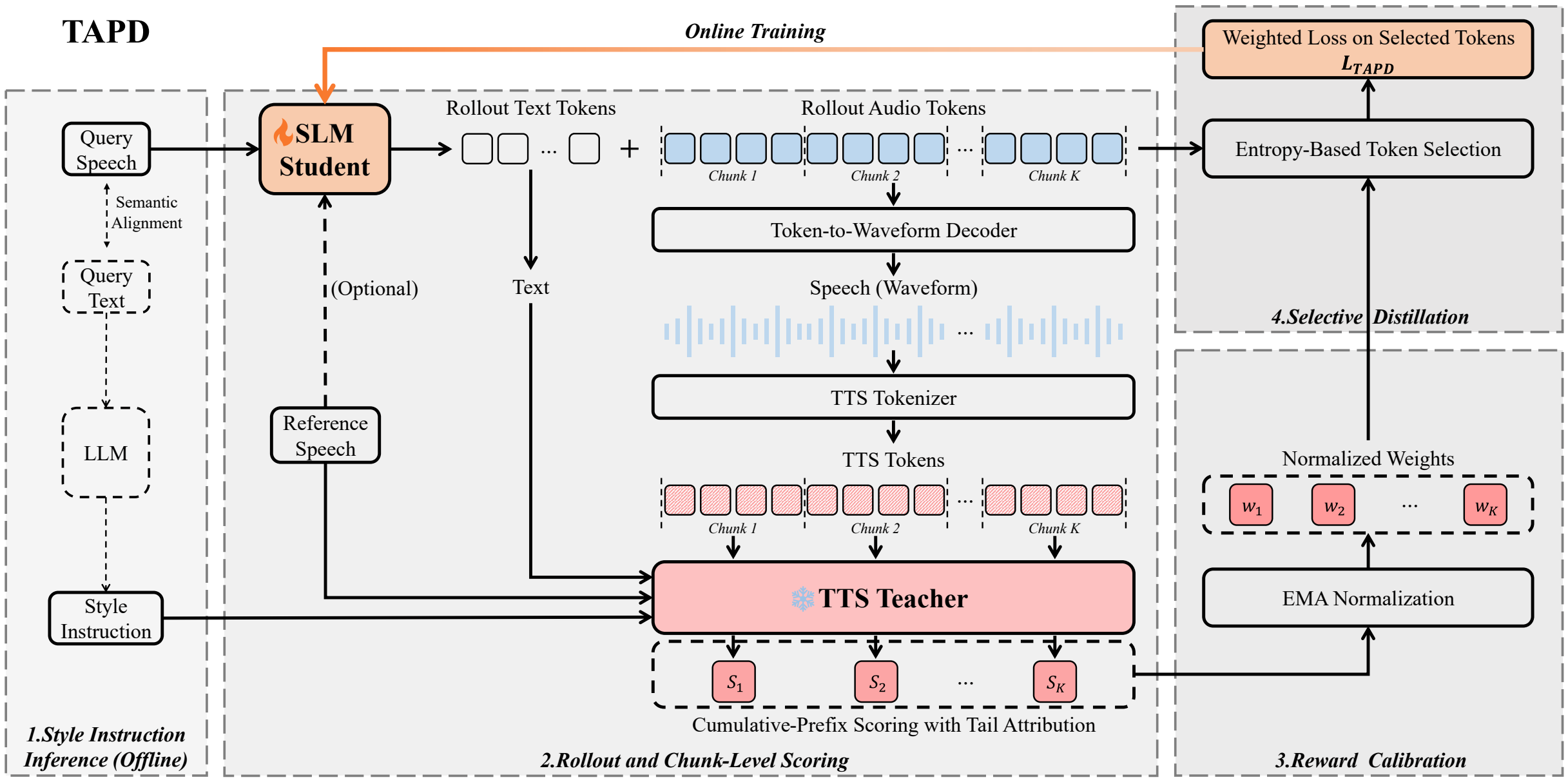}
\caption{\textbf{Overview of TAPD.}
A frozen controllable TTS teacher scores
cumulative speech prefixes generated by the student speech language
model (SLM) under teacher forcing. Tail NLLs provide EMA-normalized
chunk feedback for high-entropy audio-token positions.}
\label{fig2}
\end{figure*}

We propose \textbf{TAPD}, a \textbf{TTS-guided acoustic policy distillation} framework for improving paralinguistic generation in speech language models.
As illustrated in Figure~\ref{fig2}, the student first samples a complete response containing text and audio tokens. 
TAPD evaluates cumulative speech prefixes with a frozen controllable TTS teacher, attributes the teacher-side tail NLL to each student audio chunk, and applies the calibrated chunk weights only to high-entropy audio-token positions. 
An auxiliary SFT objective preserves generation stability.

\subsection{Style Instruction Inference}

For each training example, we obtain a style instruction describing how the response should be spoken.
Given the query text $x$ and an optional style label $\eta$, a stronger frozen LLM parser $q$ produces a TTS-compatible instruction
$z=q(x,\eta)$.
This inference is performed once during offline preprocessing, and the resulting instruction is cached for training.
The label $\eta$ is used only as an optional hint; when no label is available, $z$ is inferred directly from the query text.
The instruction may specify emotion, speaking rate, loudness, or a composite speaking style.
TAPD is not tied to a particular parser, and dataset-provided style instructions can be used directly when available.

During TAPD training, the student generates a complete response conditioned only on the original prompt $x$.
After the rollout is completed, the TTS teacher evaluates the generated speech conditioned on the cached style instruction $z$, the complete rollout text $\hat{u}$, and the reference prompt speech $r^{\mathrm{p}}$.
The same complete rollout text is used for every chunk, allowing the teacher to assess acoustic compatibility with the generated content, target voice, and intended speaking style without online LLM inference.

\subsection{Rollout and Chunk-Level Scoring}

Given a prompt $x$, the student speech LLM $\pi_\theta$ autoregressively samples a complete response
$y=(y_1,\ldots,y_T)$ until the end-of-response token is produced.
Let $\hat{u}$ denote the complete assistant text extracted from the rollout, and let
$a=(a_1,\ldots,a_N)$ denote the valid assistant audio-token sequence.
Non-assistant, non-audio, padded, masked, and post-EOS positions are excluded from acoustic policy distillation.

We partition $a$ into consecutive non-overlapping chunks
$\{C_i\}_{i=1}^{L}$ of $K$ audio tokens, merging a tail shorter than K into the preceding chunk.
Let $e_i$ denote the ending audio-token index of $C_i$, let $\mathcal{P}_i$ denote its corresponding positions in the original rollout sequence, and define the cumulative audio prefix up to the current chunk as
$a_{\leq i}=(a_1,\ldots,a_{e_i})$.
For each chunk $C_i$, the student's token-to-waveform decoder $\mathcal{D}$ converts the cumulative prefix into waveform speech:
\begin{equation}
\hat{r}_{\leq i}=\mathcal{D}(a_{\leq i}).
\label{eq:cumulative_decode}
\end{equation}
Evaluating the cumulative prefix rather than decoding $C_i$ in isolation preserves the complete preceding acoustic context.

The frozen controllable TTS teacher evaluates $\hat{r}_{\leq i}$ under teacher forcing, conditioned on the complete rollout text $\hat{u}$, reference prompt speech $r^{\mathrm{p}}$, and style instruction $z$.
Let
\[
\tilde{a}^{(i)}
=
(\tilde{a}^{(i)}_1,\ldots,\tilde{a}^{(i)}_{M_i})
=
E_{\mathcal{T}}(\hat{r}_{\leq i})
\]
denote the teacher-side acoustic target sequence extracted from the cumulative speech prefix, where $M_i$ is its length.
The teacher produces a position-wise NLL sequence over these acoustic targets.
To avoid repeatedly attributing the scores of preceding chunks, we retain only the teacher-side suffix newly introduced relative to the previous cumulative prefix:
\begin{equation}
\mathcal{A}_i
=
\{M_{i-1}+1,\ldots,M_i\},
\qquad M_0=0.
\label{eq:new_teacher_region}
\end{equation}
We associate this newly introduced suffix with the current student chunk $C_i$.
This chunk-level attribution does not require the student and teacher to share an acoustic tokenizer, token rate, or token boundaries.
Heterogeneous acoustic representations may introduce discrepancies near chunk boundaries. Supplementary Material analyzes prefix-length monotonicity and retokenization stability.

The acoustic compatibility score of chunk $C_i$ is defined as the mean teacher-forcing NLL over the newly introduced suffix:
\begin{equation}
\begin{aligned}
s_i =
-\frac{1}{|\mathcal{A}_i|}
\sum_{u\in\mathcal{A}_i}
\log p_{\mathcal{T}}\big(
&\tilde{a}^{(i)}_u
\mid
\tilde{a}^{(i)}_{<u},
r^{\mathrm{p}},
z,
\hat{u}
\big).
\end{aligned}
\label{eq:chunk_score}
\end{equation}
A lower $s_i$ serves as a proxy for greater compatibility between the speech segment introduced by chunk $C_i$ and the generated text, target voice, and requested speaking style, given the preceding acoustic context.
Earlier teacher-side positions remain available as teacher-forcing context but are excluded from the score attributed to the current chunk.

\subsection{Reward Calibration}

Raw TTS NLL scores can exhibit sample- and time-dependent scale
variation, so using them directly as optimization weights may
conflate acoustic compatibility with score scale.

We therefore calibrate chunk scores using global running statistics.
Let $\mu_t$ and $\sigma_t^2$ denote the pre-update exponential moving
average (EMA) estimates of the mean and variance of valid chunk scores
at training step $t$, respectively, with
$\sigma_t=\sqrt{\sigma_t^2}$.
We compute a normalized preference signal and apply symmetric
clipping:
\begin{equation}
\widetilde{\Delta}_i
=
\operatorname{clip}\!\left(
\frac{\mu_t-s_i}{\sigma_t+\epsilon},
-c,c
\right),
\label{eq:normalized_advantage}
\end{equation}
where $\epsilon=10^{-6}$ ensures numerical stability and $c$ limits the
influence of extreme chunk scores.

We use global EMA statistics rather than normalizing scores within
each utterance.
Utterance-wise normalization only compares chunks within the same
response and discards score differences across responses, potentially
assigning misleading preference directions when an utterance is
uniformly good or poor.
In contrast, global EMA provides a shared, temporally smoothed
empirical reference across batches and training steps.

Small deviations around the running mean may reflect teacher-score
noise rather than meaningful acoustic differences.
We therefore introduce a symmetric dead zone:
\begin{equation}
w_i
=
\widetilde{\Delta}_i\,
\mathbb{I}\!\left[
|\widetilde{\Delta}_i|\geq d
\right],
\label{eq:chunk_weight}
\end{equation}
where $d$ is the dead-zone threshold.
Clipping limits extreme updates, whereas the dead zone removes
low-confidence updates with ambiguous preference directions.
Since $s_i$ is an NLL, a below-average score yields $w_i>0$ and
reinforces the corresponding chunk, while an above-average score
yields $w_i<0$ and suppresses it.

After each optimization step, the EMA statistics are updated using
the global mean and variance of valid chunk scores in the current
batch.
The current scores are normalized using the pre-update statistics,
so they do not simultaneously define and use the same calibration
reference.

\subsection{Selective Distillation}

The TTS teacher provides chunk-level acoustic feedback rather than exact supervision for every student token.
Applying the same chunk weight densely to all audio-token positions may therefore overwrite reliable decisions already learned through supervised fine-tuning.
TAPD instead applies acoustic policy distillation only to uncertain student audio-token positions.

Motivated by entropy-based token selection in knowledge distillation
\cite{DBLP:conf/eacl/VuKDLSL26,DBLP:journals/corr/abs-2604-14084}, TAPD applies chunk-level acoustic feedback
only to uncertain student audio-token decisions. For each valid audio position $t$, let
$\bar{p}_\theta(\cdot\mid y_{<t},x)$ denote the student's next-token distribution restricted and renormalized over the audio vocabulary $\mathcal{V}_{\mathrm{audio}}$.
We compute
\begin{equation}
H_t =
-\sum_{v\in\mathcal{V}_{\mathrm{audio}}}
\bar{p}_\theta(v\mid y_{<t},x)
\log
\bar{p}_\theta(v\mid y_{<t},x).
\label{eq:audio_entropy}
\end{equation}
Within each completed rollout, we retain the top $\rho$ fraction of valid assistant audio-token positions ranked by $H_t$ and denote this set by $\mathcal{S}$.
The selection set is treated as fixed during backpropagation.

For chunk $C_i$, the selective token-level loss is
\begin{equation}
\ell_i^{\mathrm{sel}}
=
-\frac{1}{|\mathcal{P}_i\cap\mathcal{S}|}
\sum_{t\in\mathcal{P}_i\cap\mathcal{S}}
\log \pi_\theta(y_t\mid y_{<t},x).
\label{eq:selective_loss}
\end{equation}
Averaging within each chunk prevents its contribution from depending on the number of selected positions.
Chunks with no selected token positions are excluded.

Let $\mathcal{I}$ contain chunks with
$|\mathcal{P}_i\cap\mathcal{S}|>0$, including those with $w_i=0$.
The TAPD loss is
\begin{equation}
\mathcal{L}_{\mathrm{TAPD}}
=
\frac{1}{|\mathcal{I}|}
\sum_{i\in\mathcal{I}}
\operatorname{sg}(w_i)\,
\ell_i^{\mathrm{sel}},
\label{eq:tapd_loss}
\end{equation}
where $\operatorname{sg}(\cdot)$ denotes stop-gradient.
If $\mathcal{I}$ is empty, we set $\mathcal{L}_{\mathrm{TAPD}}=0$.
Minimizing this objective reinforces low-NLL chunks preferred by the teacher and suppresses high-NLL chunks, while modifying only uncertain audio-token decisions.

We jointly optimize TAPD with a standard SFT loss weighted by $\lambda_{\mathrm{SFT}}$ to preserve instruction following and generation stability.
The sampled rollout, teacher-derived chunk weights, and entropy-based selection set are treated as fixed for the current update.
Algorithm~\ref{alg:TAPD} summarizes the complete training procedure.

\begin{algorithm}[t]
\caption{TAPD Training}
\label{alg:TAPD}
\small
\begin{algorithmic}[1]
\REQUIRE Student speech LLM $\pi_\theta$; frozen TTS teacher
$\mathcal{T}$; token-to-waveform decoder $\mathcal{D}$; training data
$\mathcal{D}_{\mathrm{sft}}$ with cached style instructions;
hyperparameters $K,\rho,\beta,c,d,\lambda_{\mathrm{SFT}}$;
EMA statistics $(\mu,\sigma^2)$

\FOR{each training step}
    \STATE Sample $(x,r^{\mathrm{p}},z,y^\star)
    \sim\mathcal{D}_{\mathrm{sft}}$
    \STATE Sample a complete rollout
    $y\sim\pi_\theta(\cdot\mid x)$ and extract
    $(\hat{u},a)$
    \STATE Partition $a$ into chunks $\{C_i\}$ using the
    tail-merging rule; set $M_0\gets0$

    \FOR{each chunk $C_i$ in sequence}
        \STATE Decode the cumulative prefix
        $\hat{r}_{\le i}=\mathcal{D}(a_{\le e_i})$
        \STATE Teacher-force $\mathcal{T}$ on $\hat{r}_{\le i}$
        conditioned on $(r^{\mathrm{p}},z,\hat{u})$, obtaining
        NLLs $\{\ell_u^{\mathcal T}\}_{u=1}^{M_i}$
        \STATE $s_i\gets
        \frac{1}{M_i-M_{i-1}}
        \sum_{u=M_{i-1}+1}^{M_i}\ell_u^{\mathcal T}$
    \ENDFOR

    \STATE Convert $\{s_i\}$ into weights $\{w_i\}$ using
    pre-update EMA statistics, clipping $c$, and dead zone $d$
    \STATE Select the top-$\rho$ valid audio positions
    $\mathcal{S}$ according to audio-vocabulary entropy
    \STATE Compute $\mathcal{L}_{\mathrm{TAPD}}$ on $\mathcal{S}$
    and $\mathcal{L}_{\mathrm{SFT}}$ on $y^\star$
    \STATE Update $\theta$ using
    $\mathcal{L}_{\mathrm{TAPD}}
    +\lambda_{\mathrm{SFT}}\mathcal{L}_{\mathrm{SFT}}$
    \STATE Update $(\mu,\sigma^2)$ with decay $\beta$ using
    the global moments of valid chunk scores
\ENDFOR
\end{algorithmic}
\end{algorithm}

\section{Experiments}

\subsection{Experimental Setup}
\label{sec:experimental_setup}

\paragraph{Dataset and Evaluation Metrics.}
We conduct all experiments on the UltraVoice benchmark~\cite{DBLP:journals/corr/abs-2510-22588}, which covers multiple dimensions of paralinguistic speech control.
We use its four paralinguistic subsets: \textit{emotion}, \textit{speed}, \textit{volume}, and \textit{composite}, where the composite subset requires the model to satisfy multiple speaking-style constraints simultaneously.
We construct a balanced training set of 10{,}000 examples by sampling 2{,}500 examples from each subset.
For evaluation, we use the complete test set of each of the four subsets without additional sampling.
Following the UltraVoice protocol~\cite{DBLP:journals/corr/abs-2510-22588}, we use Gemini-2.5-Flash~\cite{DBLP:journals/corr/abs-2507-06261} with the same evaluation prompts as the original benchmark to assess every generated response.
We report an automatic mean opinion score (MOS) and instruction-following rate (IFR) for each subset.
MOS evaluates the overall quality of the generated spoken response, including response content and paralinguistic realization, whereas IFR measures whether the generated speech satisfies the requested paralinguistic constraints, independently of response correctness.
Because this work focuses on paralinguistic controllability, we place greater emphasis on IFR.
On 100 class-balanced samples annotated by two human raters, Gemini-based IFR achieved 87.5\% mean agreement (94\%/81\% for predicted IFR $=0/1$; see Supplementary Material).

\paragraph{Student and Teachers.}
We use the 7B-parameter Step-Audio 2 \cite{DBLP:journals/corr/abs-2507-16632} as the student speech language model and evaluate TAPD with two controllable TTS teachers: CosyVoice2 \cite{DBLP:journals/corr/abs-2412-10117} and CosyVoice3 \cite{DBLP:journals/corr/abs-2505-17589}, each containing approximately 0.5B parameters.
Both TTS teachers remain frozen throughout training.
For each teacher, we use Qwen3-32B \cite{DBLP:journals/corr/abs-2505-09388} as the frozen LLM parser to infer style instructions once during offline preprocessing, which are then cached for training.
We construct a matched experimental setting for each TTS teacher, allowing us to examine whether TAPD consistently improves the same student model using different compact acoustic teachers.

\paragraph{Baselines.}
We compare TAPD with the following baselines.
\textbf{TTS Teacher} directly synthesizes speech using the corresponding controllable TTS model and serves as a reference for its instruction-conditioned synthesis capability.
\textbf{Original Student} denotes the pretrained Step-Audio 2 model without additional task-specific training.
\textbf{Controllable-TTS SFT} fine-tunes the student on speech synthesized with the same offline style-instruction inference and controllable-TTS pipeline employed by TAPD, representing offline imitation of instruction-conditioned TTS data.
\textbf{Sentence-Level TTS-NLL} evaluates each complete student-generated utterance with the TTS teacher, converts the resulting teacher-forcing NLL into a single sentence-level reward, and normalizes it using sentence-level EMA statistics.
\textbf{TTS-GRPO} uses the same sentence-level TTS-derived reward and performs group-relative policy optimization (GRPO) \cite{DBLP:journals/corr/abs-2402-03300} with three student rollouts per prompt.
We do not further increase the rollout group size because additional rollouts substantially increase training time beyond that of TAPD, resulting in a computationally imbalanced comparison.
Finally, \textbf{TAPD} provides localized chunk-level feedback by evaluating cumulative speech prefixes with the controllable TTS teacher and assigning the resulting teacher-forcing acoustic scores to the corresponding regions of the student's on-policy speech trajectory.
All trainable methods use the same balanced 10{,}000-example training set.
Detailed implementations of the sentence-level and TTS-GRPO baselines are provided in Supplementary Material.

\paragraph{Training Details.}
Across all TAPD main experiments, we use a chunk size of $K=12$, an entropy-based token selection ratio of $\rho=0.5$, and an EMA decay factor of $\beta=0.95$. We use symmetric positive and negative clipping thresholds of $c=1.0$ and set the dead-zone threshold to $d=0.5$. The same TAPD hyperparameters are used for both the CosyVoice2 and CosyVoice3 teacher settings without teacher-specific tuning. For a fair comparison, all trainable student methods use matched optimizer settings, numbers of training updates, and random seeds unless otherwise specified. Under the main setting, each response contains 18 audio chunks on average. The general scoring path evaluates each cumulative prefix separately with the 0.5B TTS teacher; its costs are reported in Supplementary Material. Due to the main-text space limit, additional implementation details and analyses are provided in Supplementary Material, including prompt templates, data preprocessing, optimization and generation configurations, complete hyperparameter settings, baseline implementations, hyperparameter analyses, and a detailed analysis of training efficiency and computational cost.

\begin{table*}[t]
\centering

\small
\setlength{\tabcolsep}{3.8pt}
\renewcommand{\arraystretch}{1.10}

\begin{tabular}{lcccccccccc}
\toprule
& \multicolumn{5}{c}{\textbf{MOS} $\uparrow$}
& \multicolumn{5}{c}{\textbf{IFR (\%)} $\uparrow$} \\
\cmidrule(lr){2-6}
\cmidrule(lr){7-11}
\textbf{Method}
& \textbf{Emotion}
& \textbf{Speed}
& \textbf{Volume}
& \textbf{Composite}
& \textbf{Avg.}
& \textbf{Emotion}
& \textbf{Speed}
& \textbf{Volume}
& \textbf{Composite}
& \textbf{Avg.} \\
\midrule

\rowcolor{gray!15}
\multicolumn{11}{c}{\textbf{Acoustic Teacher: CosyVoice3}} \\
\specialrule{0.6pt}{0pt}{2pt}

TTS Teacher
& 4.576 & 4.827 & 4.717 & 4.570 & 4.672
& 72.71 & 93.00 & 78.00 & 79.00 & 80.68 \\

Original Student
& 3.024 & 3.990 & 3.878 & 3.780 & 3.668
& 29.71 & 82.67 & 70.33 & 57.00 & 59.93 \\

Controllable-TTS SFT
& 3.592 & 4.473 & 4.362 & 4.190 & 4.154
& 41.14 & 83.33 & 68.00 & 61.00 & 63.37 \\

Sentence-Level
& 3.531 & 4.348 & 4.043 & 4.135 & 4.015
& 42.43 & 84.67 & 69.33 & 64.00 & 65.11 \\

TTS-GRPO (G=3)
& 3.537 & 4.437 & 4.132 & 4.195 & 4.075
& 40.86 & 85.33 & 70.33 & 56.00 & 63.13 \\

\textbf{TAPD (Ours)}
& \textbf{3.748} & \textbf{4.532} & \textbf{4.455} & \textbf{4.205} & \textbf{4.235}
& \textbf{43.29} & \textbf{86.67} & \textbf{73.67} & \textbf{72.00} & \textbf{68.90} \\

\midrule

\rowcolor{gray!15}
\multicolumn{11}{c}{\textbf{Acoustic Teacher: CosyVoice2}} \\
\specialrule{0.6pt}{0pt}{2pt}

TTS Teacher
& 4.491 & 4.785 & 4.538 & 4.590 & 4.601
& 69.57 & 86.67 & 77.33 & 69.00 & 75.64 \\

Original Student
& 3.024 & 3.990 & 3.878 & 3.780 & 3.668
& 29.71 & 82.67 & 70.33 & 57.00 & 59.93 \\

Controllable-TTS SFT
& 3.628 & 4.387 & 4.200 & 4.170 & 4.096
& 42.43 & 83.33 & 70.00 & 59.00 & 63.69 \\

Sentence-Level
& 3.601 & 4.410 & 4.133 & 4.030 & 4.044
& 42.57 & 84.33 & 70.67 & 64.00 & 65.39 \\

TTS-GRPO (G=3)
& 3.646 & 4.390 & 4.256 & \textbf{4.195} & 4.122
& 41.86 & 84.33 & 68.00 & 63.00 & 64.30 \\

\textbf{TAPD (Ours)}
& \textbf{3.655} & \textbf{4.502} & \textbf{4.283} & 4.140 & \textbf{4.145}
& \textbf{43.00} & \textbf{84.67} & \textbf{75.33} & \textbf{72.00} & \textbf{68.75} \\

\bottomrule
\end{tabular}

\caption{
Main results on UltraVoice.
The TTS Teacher serves as a reference.
Avg. denotes the unweighted average over the four subsets.
}
\label{tab:main_results}
\end{table*}

\subsection{TTS Score Validation}
\label{sec:tts_score_validation}

Before using TTS likelihood as acoustic feedback, we examine the granularity at which teacher-forcing negative log-likelihood (NLL) provides reliable supervision.
We evaluate both CosyVoice3 and CosyVoice2; complete protocols and results are provided in Supplementary Material.
We treat lower NLL as a proxy for compatibility with the conditioned text, reference voice, and speaking style.

We first score the same utterance under matched and conflicting paralinguistic instructions, keeping the speech, text, and reference speaker fixed.
Sentence-level average NLL provides limited discrimination, with pairwise accuracy around 60--70\% and weaker performance on some speed and volume conditions.
In contrast, controlled local perturbations, including simulated pronunciation errors and localized changes in speaking rate and volume, are distinguished from matched clean regions with over 95\% pairwise accuracy.
Although pronunciation and paralinguistic errors represent different linguistic phenomena, both appear to the TTS teacher as localized acoustic mismatches with the conditioned realization.

\begin{figure}[t]
    \centering
    \includegraphics[width=0.9\linewidth]{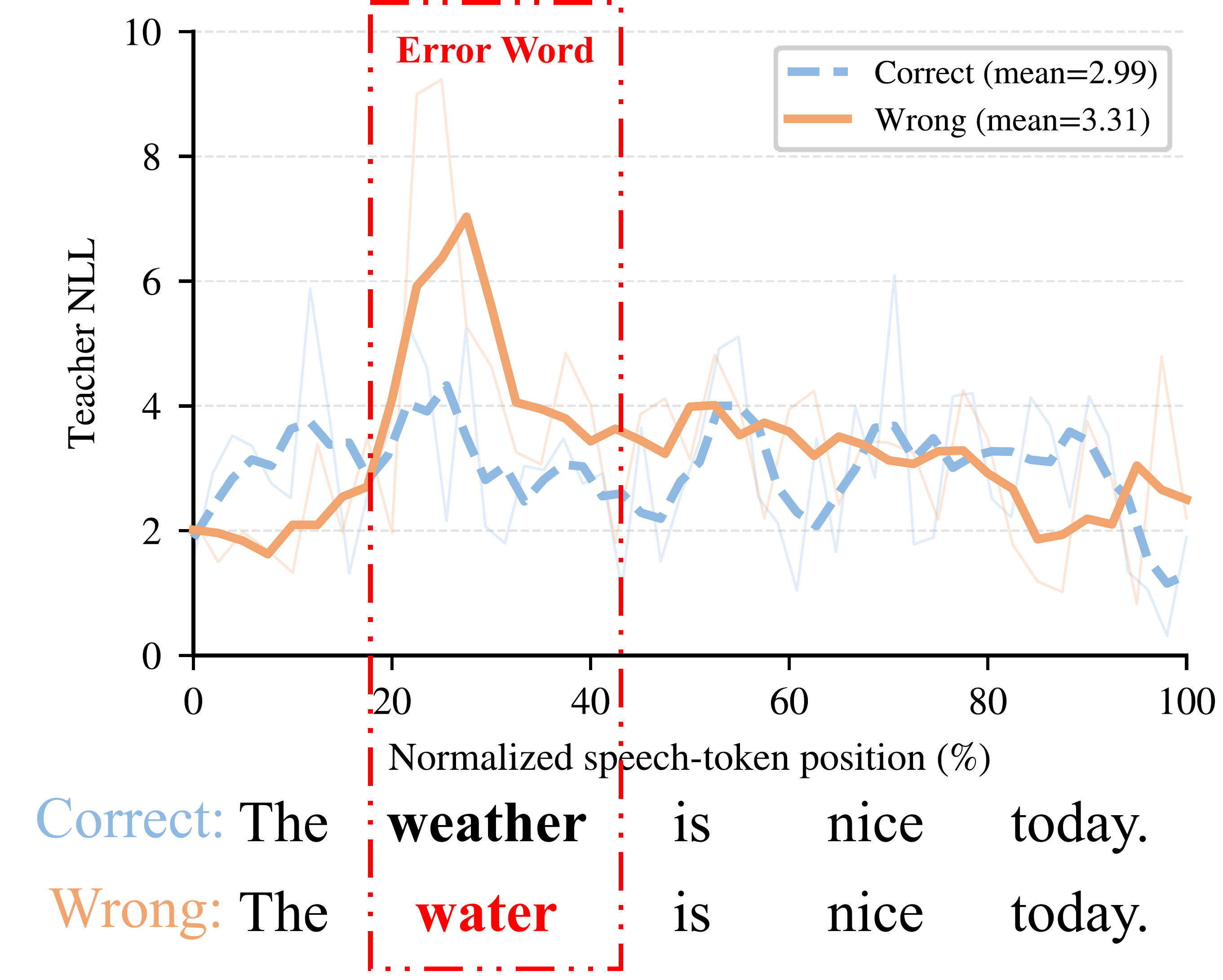}
    \caption{
    Token-level teacher NLL for a simulated pronunciation error, showing a localized increase around the mismatch.
    }
    \label{fig:wrong_word_case}
\end{figure}

Figure~\ref{fig:wrong_word_case} illustrates a representative case in which the intended word ``weather'' is realized as ``water''.
The teacher NLL increases clearly around the mismatched region while remaining relatively stable elsewhere.
Complete experiments and analyses of both pronunciation perturbations and localized paralinguistic perturbations are provided in Supplementary Material.
These results motivate localized TTS-likelihood feedback, as sentence-level averaging can dilute regional acoustic errors.

\subsection{Main Results}
\label{sec:main_results}

Table~\ref{tab:main_results} reports the results on UltraVoice with CosyVoice3 and CosyVoice2 as acoustic teachers.
The TTS Teacher represents speech synthesized directly by each controllable TTS model, while all other methods use the same Step-Audio 2 student.
Since MOS also depends on response content and general speech quality, we primarily analyze IFR, which more directly evaluates whether the requested paralinguistic constraints are satisfied.

\paragraph{Comparison with Baselines.}
Under the CosyVoice3 setting, the original student obtains an average IFR of 59.93\%, substantially below the TTS teacher at 80.68\%.
Controllable-TTS SFT improves IFR to 63.37\%, confirming the value of instruction-conditioned synthetic speech, but still leaves a 17.31-point teacher--student gap.
Sentence-Level TTS-NLL reaches 65.11\%, while TTS-GRPO obtains 63.13\%, suggesting that a single utterance-level reward cannot adequately localize acoustic errors.

TAPD achieves the strongest student-side performance, reaching an average IFR of 68.90\%.
It outperforms Controllable-TTS SFT, Sentence-Level TTS-NLL, and TTS-GRPO by 5.53, 3.79, and 5.77 percentage points, respectively, while also obtaining the highest average MOS of 4.235.
These results demonstrate that cumulative-prefix scoring and chunk-level credit assignment provide more effective acoustic supervision than either fixed synthetic references or sentence-level rewards.

\paragraph{Dimension-Level Improvements.}
Compared with Controllable-TTS SFT, TAPD improves IFR by 2.15 points on emotion, 3.34 on speed, 5.67 on volume, and 11.00 on composite styles.
The largest composite-style gain suggests that localized on-policy feedback is especially valuable when multiple acoustic constraints must be maintained throughout autoregressive generation.
Moreover, relative to the original student, TAPD improves emotion and composite IFR by 13.58 and 15.00 points, respectively, the dimensions with the largest teacher--student capability gaps.

\paragraph{Consistency across Teachers.}
The CosyVoice2 results show the same trend.
TAPD reaches an average IFR of 68.75\%, exceeding Controllable-TTS SFT, Sentence-Level TTS-NLL, and TTS-GRPO by 5.06, 3.36, and 4.45 points.
The consistent improvements under two teachers, using the same hyperparameters, demonstrate that TAPD is not specific to a particular controllable TTS model.

\subsection{Ablation Studies}
\label{sec:ablation}

\begin{figure}[t]
    \centering
    \includegraphics[width=0.9\linewidth]{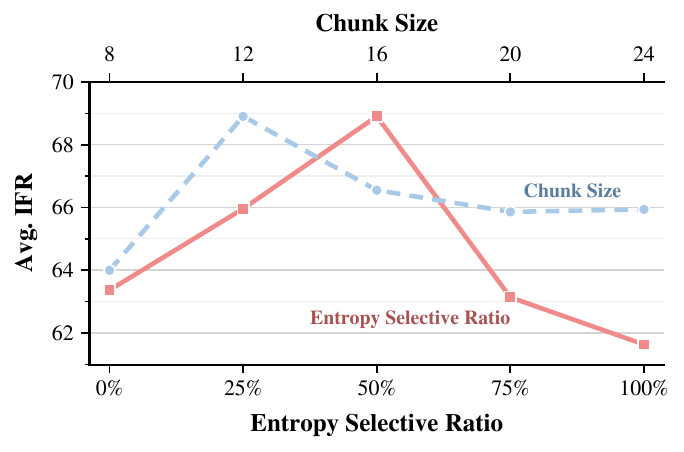}
    \caption{Average IFR under different chunk sizes and entropy-selection ratios.}
    \label{fig:hyperparameter_sensitivity}
\end{figure}

We conduct ablation studies with CosyVoice3 as the acoustic teacher.
Table~\ref{tab:ablation} examines the main components of TAPD, while Figure~\ref{fig:hyperparameter_sensitivity} studies the effects of chunk size and entropy-based token selection.
The Controllable-TTS SFT and Sentence-Level baselines in Table~\ref{tab:main_results} additionally provide comparisons without the TTS reward and with response-level reward granularity, respectively.
Since IFR more directly reflects paralinguistic instruction following, we primarily analyze IFR below.

\begin{table}[t]
\centering
\small
\setlength{\tabcolsep}{5.0pt}
\renewcommand{\arraystretch}{1.08}
\begin{tabular}{lcc}
\toprule
\textbf{Method}
& \textbf{MOS Avg.} $\uparrow$
& \textbf{IFR Avg.} $\uparrow$ \\
\midrule
\textbf{TAPD (Ours)}
& \textbf{4.235} & \textbf{68.90} \\
\midrule
w/o Style Instruction
& 4.102 & 66.45 \\
w/o Dead Zone
& 4.122 & 66.62 \\
Utterance-wise Normalization
& 4.119 & 65.33 \\
Random Token Selection
& 3.526 & 62.21 \\
\bottomrule
\end{tabular}
\caption{Component ablations with CosyVoice3 as the acoustic teacher.}
\label{tab:ablation}
\end{table}

\paragraph{Acoustic Teacher Signal and Conditioning.}
The Controllable-TTS SFT baseline in Table~\ref{tab:main_results},
which removes on-policy TTS feedback while retaining the same
offline supervision, achieves an average IFR of 63.37\%, compared
with 68.90\% for TAPD. This confirms that the gains do not arise
solely from additional training on controllable-TTS data.
For w/o Style Instruction, we replace each target-specific instruction with a fixed neutral instruction, reducing IFR by 2.45 percentage points to 66.45\%. This shows that effective TTS feedback requires
conditioning on the requested paralinguistic style rather than a
generic speaking style.

\paragraph{Reward Calibration.}
Removing the dead zone decreases IFR from 68.90\% to 66.62\%, indicating that updates from near-baseline chunks may be dominated by score noise and should be suppressed.
Replacing global EMA normalization with utterance-wise normalization causes a larger drop to 65.33\%.
Sentence-internal statistics judge each chunk only relative to other chunks in the same utterance, discarding acoustic-quality differences across utterances and potentially assigning opposing weights even when all chunks are uniformly good or poor.
Global EMA statistics instead provide a more stable reference across batches and training steps.

\paragraph{Entropy-Based Token Selection.}
Randomly selecting the same 50\% of audio-token positions reduces IFR to 62.21\%, demonstrating that the improvement does not arise merely from optimizing fewer positions.
In Figure~\ref{fig:hyperparameter_sensitivity}, the 0\% setting applies no TAPD update and is equivalent to the Controllable-TTS SFT baseline, whereas the 100\% setting corresponds to dense distillation over all valid audio tokens.
Increasing the entropy-selection ratio from 0\% to 25\% and 50\% improves IFR from 63.37\% to 65.96\% and 68.90\%, respectively.
However, IFR decreases to 63.14\% and 61.64\% at ratios of 75\% and 100\%.
Updating too many positions includes highly confident audio-token decisions, which can overwrite reliable generation behavior and amplify coarse reward noise.
A ratio of 50\% therefore provides the best balance between correcting uncertain decisions and preserving confident ones.

\paragraph{Chunk Granularity.}
A chunk size of 12 achieves the highest IFR of 68.90\%.
Reducing it to 8 lowers IFR to 64.00\%, as very short waveform segments contain insufficient acoustic context for reliable TTS evaluation.
We do not test smaller values because such short prefixes carry little stable paralinguistic information and are not reliably supported by the token-to-waveform decoder.
Increasing the chunk size to 16--24 also reduces IFR to 65.86--66.55\%, since coarser rewards may dilute localized acoustic errors and introduce less accurate credit assignment.
This trend is consistent with the Sentence-Level baseline in Table~\ref{tab:main_results}, which further expands the scoring granularity to the complete response and obtains only 65.11\% IFR.
These results support an intermediate chunk granularity that provides sufficient acoustic context while retaining localized supervision.

\section{Conclusion}

We presented TAPD, a TTS-guided acoustic policy distillation framework that uses a frozen controllable TTS model to provide chunk-level feedback on student-generated speech.
By combining cumulative-prefix scoring, tail attribution, and entropy-based token selection, TAPD enables heterogeneous acoustic distillation without shared tokenization or teacher logits.
Experiments on four UltraVoice subsets with two TTS teachers show that TAPD consistently achieves the best average IFR among student methods, outperforming matched controllable-TTS SFT by over five percentage points.
These results demonstrate the potential of controllable TTS models as on-policy acoustic teachers for speech LLM post-training.

\clearpage
\bibliography{aaai2027}

\clearpage
\appendix
\setcounter{equation}{0}
\setcounter{table}{0}
\setcounter{figure}{0}
\renewcommand{\theequation}{A\arabic{equation}}
\renewcommand{\thetable}{A\arabic{table}}
\renewcommand{\thefigure}{A\arabic{figure}}
\section*{Supplementary Material}
\section{Experimental Environment and Shared Training Configuration}
\label{app:sec:experimental_setup}

This section describes the hardware environment and the training
configuration shared by TAPD and all trainable baselines. Unless otherwise
specified, all post-training experiments use the same student initialization,
training data, optimization configuration, distributed-training strategy,
and random seed. Method-specific settings for TAPD and individual baselines
are provided in subsequent sections.

\subsection{Hardware Environment}
\label{app:sec:hardware_environment}

Each experiment used either eight NVIDIA A800 GPUs with 80\,GB of memory
per GPU or eight NVIDIA A100 GPUs with 80\,GB of memory per GPU. Each server
was equipped with two Intel Xeon Platinum 8358P processors running at
2.60\,GHz. Each processor contains 32 physical CPU cores, resulting in a
total of 64 physical CPU cores per server. The servers were additionally
equipped with 1\,TB of system memory.

\subsection{Software Environment}
\label{app:sec:software_environment}

All experiments were conducted under Ubuntu 20.04.5 LTS. Distributed
training uses DeepSpeed ZeRO Stage~2 with BF16 precision; the shared
training settings are listed in
Table~\ref{app:tab:shared_training_configuration}. Exact software-package
versions are not included in this preprint. The environment specification
will accompany the public code release described in
Section~\ref{app:sec:code_availability}.

\subsection{Shared Training Configuration}
\label{app:sec:shared_training_configuration}

For all post-training experiments, we use Step-Audio-2-mini, the mini variant
of Step-Audio~2, as the student initialization. All methods are trained on the
same set of 10,000 utterances. The audio encoder and adapter are frozen during
post-training, while all remaining student parameters are updated.

Unless otherwise stated, TAPD and all trainable baselines use identical
optimization settings, distributed-training configurations, and random seeds.
Training is performed with DeepSpeed ZeRO Stage~2, which partitions optimizer
states and gradients across the eight distributed worker processes. The
complete shared configuration is summarized in
Table~\ref{app:tab:shared_training_configuration}.

\begin{table*}[tp]
\centering
\small
\setlength{\tabcolsep}{5pt}
\renewcommand{\arraystretch}{1.12}
\caption{Training configuration shared by TAPD and all trainable baselines.
Step-Audio-2-mini is the mini variant of Step-Audio~2. The effective batch
size is computed as the product of the per-device micro-batch size, the
number of distributed processes, and the number of gradient-accumulation
steps.}
\label{app:tab:shared_training_configuration}
\begin{tabularx}{\textwidth}{
    >{\raggedright\arraybackslash}p{0.23\textwidth}
    >{\raggedright\arraybackslash}p{0.22\textwidth}
    >{\raggedright\arraybackslash}X
}
\toprule
\textbf{Setting} & \textbf{Value} & \textbf{Description} \\
\midrule

Student model
& Step-Audio-2-mini
& The mini variant of Step-Audio~2, used to initialize the student in all
post-training experiments. \\

Training data
& 10,000 utterances
& Shared training set used by all trainable methods. \\

Optimizer
& AdamW
& Optimizer coefficients are $\beta_1=0.9$ and $\beta_2=0.999$. \\

Learning rate
& $5\times10^{-6}$
& Peak learning rate reached after warmup and kept constant thereafter. \\

Learning-rate schedule
& Constant with warmup
& Linear learning-rate warmup followed by a constant learning rate. \\

Warmup ratio
& 0.1
& The first 10\% of the optimization steps are used for linear warmup. \\

Weight decay
& 0.01
& Weight decay is not applied to bias or normalization parameters. \\

Training epochs
& 1
& Each method is trained for one pass over the training set. \\

Number of GPUs
& 8
& Training uses eight distributed worker processes, one per GPU. \\

Micro-batch size
& 1 per GPU
& Per-device batch size for each forward and backward pass. \\

Gradient accumulation
& 8
& Number of micro-steps accumulated before each optimizer update. \\

Effective batch size
& 64 utterances
& Computed as $1 \times 8 \times 8$ utterances per optimizer update. \\

Numerical precision
& BF16
& Mixed-precision training using the bfloat16 numerical format. \\

Distributed strategy
& DeepSpeed ZeRO Stage 2
& Optimizer states and gradients are partitioned across the eight
distributed worker processes. \\

Maximum sequence length
& 16,384 tokens
& Maximum student sequence length used during post-training. \\

Gradient clipping
& 1.0
& Maximum global gradient norm. \\

Gradient checkpointing
& Disabled
& Activations are retained without recomputation during backpropagation. \\

Frozen modules
& Audio encoder and adapter
& These modules remain frozen; all remaining student parameters are updated. \\

Random seed
& 42
& Shared random seed used across all training experiments. \\

\bottomrule
\end{tabularx}
\end{table*}

\section{Exact TAPD Implementation}
\label{app:sec:exact_tapd_implementation}

This section provides the implementation details specific to TAPD.
The shared optimization and distributed-training configuration is
described in Section~\ref{app:sec:experimental_setup}, while the
method-specific configurations of the baselines and the rollout
generation settings are provided in subsequent sections.

\subsection{TAPD-Specific Configuration}
\label{app:sec:tapd_specific_configuration}

Table~\ref{app:tab:tapd_specific_settings} summarizes the configurations
used in all TAPD experiments. The same TAPD hyperparameters are used
with both CosyVoice2 and CosyVoice3, without teacher-specific tuning.

\begin{table*}[tp]
\centering
\small
\setlength{\tabcolsep}{5pt}
\renewcommand{\arraystretch}{1.12}
\caption{TAPD-specific configurations used in the main experiments.
The two TTS teachers are evaluated in separate matched experimental
settings. Shared student-training settings are reported in
Table~\ref{app:tab:shared_training_configuration}.}
\label{app:tab:tapd_specific_settings}
\begin{tabularx}{\textwidth}{
    >{\raggedright\arraybackslash}p{0.23\textwidth}
    >{\raggedright\arraybackslash}p{0.22\textwidth}
    >{\raggedright\arraybackslash}X
}
\toprule
\textbf{Configuration}
& \textbf{Value}
& \textbf{Description} \\
\midrule

TTS teacher
& CosyVoice2-0.5B or CosyVoice3-0.5B
& Each teacher is evaluated in a separate experimental setting and
remains frozen throughout student post-training. \\

Style-instruction parser
& Qwen3-32B
& The frozen parser converts the query and, when available, its style
label into a TTS-compatible style instruction during offline
preprocessing. \\

Instruction caching
& Enabled
& Style instructions are inferred once and cached; no additional LLM
inference is performed during TAPD training. \\

Student chunk size $K$
& 12 audio tokens
& This intermediate granularity provides sufficient acoustic context
for teacher evaluation while retaining localized credit assignment. \\

Tail-chunk handling
& Merge with the preceding chunk
& A final chunk containing fewer than $K$ audio tokens is merged into
the preceding chunk rather than scored as an isolated short segment. \\

Skipped prefix chunks
& 1
& The first student audio chunk is excluded from reward calibration and
the TAPD loss because its teacher NLL is frequently dominated by
short-prefix and heterogeneous-tokenization effects. \\

Teacher scoring mode
& Cumulative-prefix scoring
& The teacher evaluates the waveform decoded from the complete student
audio prefix up to each chunk boundary. \\

Teacher conditioning
& Rollout text, reference speech, and style instruction
& Every cumulative prefix is scored using the same complete rollout
text, reference prompt speech, and cached style instruction. \\

Chunk score
& Mean tail-region teacher NLL
& Only teacher tokens newly introduced relative to the preceding
cumulative prefix are used to compute the score assigned to the current
student chunk. \\

Prefix evaluation
& General cumulative-prefix scoring
& Each cumulative waveform is independently retokenized and teacher-scored.
Reusing cached style instructions does not imply incremental teacher
scoring or reuse of teacher acoustic KV states. \\

Reward reference
& Global chunk-score EMA
& Valid chunk scores are calibrated using running statistics shared
across utterances, batches, and training steps. \\

EMA decay $\beta$
& 0.95
& The running statistics are smoothed sufficiently to reduce
batch-level score fluctuations while remaining adaptive during
training. \\

Symmetric clipping $c$
& 1.0
& Normalized positive and negative preference weights are clipped to
$[-1,1]$ to limit the influence of extreme teacher scores. \\

Reward dead zone $d$
& 0.5
& Normalized weights with magnitude below $0.5$ are discarded to avoid
updates driven by small and potentially noisy score deviations. \\

Token-selection criterion
& Within-rollout audio-token entropy
& Valid assistant audio-token positions are ranked by the entropy of
the student distribution restricted and renormalized over the audio
vocabulary. \\

Selected-token ratio $\rho$
& 0.5
& TAPD updates the highest-entropy 50\% of valid assistant audio-token
positions and leaves lower-entropy decisions unchanged. \\

TAPD loss coefficient
& 0.5
& The selective TAPD objective is multiplied by $0.5$ before being
combined with the auxiliary SFT objective. \\

Auxiliary SFT data
& Matched controllable-TTS synthetic responses
& Ground-truth response text is synthesized using the TTS teacher
corresponding to each experimental setting. \\

SFT supervision
& Complete synthetic assistant response
& The loss supervises valid assistant text, acoustic, and output
special-token positions generated from the original query--response
pair. \\

\bottomrule
\end{tabularx}
\end{table*}

\subsection{Style Instruction and Teacher Conditioning}
\label{app:sec:tapd_teacher_conditioning}

For each training example, a frozen Qwen3-32B parser produces a
TTS-compatible style instruction from the query text and, when
available, the associated style label. This preprocessing step is
performed only once, and the resulting instruction is cached for all
subsequent training epochs.

During TAPD training, the student generates a complete response
conditioned only on the original prompt. Let $\hat{u}$ denote the
assistant text extracted from this rollout, and let $r^{p}$ denote the
reference prompt speech. The frozen TTS teacher evaluates every
cumulative student-speech prefix conditioned on the same tuple
$(\hat{u}, r^{p}, z)$, where $z$ is the cached style instruction.
Using the complete rollout text for all chunk boundaries ensures that
changes in the teacher score primarily reflect the acoustic realization
of the newly appended speech region rather than changes in textual
conditioning.

\subsubsection{Training-Time Rollout Settings}
\label{app:sec:rollout_generation_settings}

Training rollouts use the same sampling hyperparameters as evaluation,
following the official Step-Audio~2 inference recipe:\footnote{\url{https://github.com/stepfun-ai/Step-Audio2}}
stochastic sampling (\texttt{do\_sample=true}), temperature $0.7$,
top-$p=0.9$, and repetition penalty $1.05$. Top-$k$ is left at the
\texttt{GenerationConfig} default of $50$. For training efficiency, each
rollout is capped at 300 newly generated multimodal tokens, whereas
evaluation allows 800. Generation stops when the model emits
\texttt{<|EOT|>} or reaches the token cap, whichever occurs first; no
separate audio-duration limit is imposed.

The sampling top-$k$ setting is distinct from entropy-based token
selection, which uses the complete audio vocabulary as defined in
Section~\ref{app:sec:tapd_selective_distillation}. TAPD and the
sentence-level method use one rollout per input, while TTS-GRPO samples
$G=3$ responses per input. Shared training settings are reported in
Table~\ref{app:tab:shared_training_configuration}; the evaluation
configuration is summarized in
Section~\ref{app:sec:evaluation_inference_configuration}.

\subsection{Chunk Construction and Cumulative-Prefix Scoring}
\label{app:sec:tapd_chunk_scoring}

Let
\[
a=(a_1,\ldots,a_N)
\]
denote the valid assistant audio-token sequence extracted from a
student rollout. Non-assistant, non-audio, padded, masked, and
post-EOS positions are excluded. We divide $a$ into consecutive,
non-overlapping chunks
\[
\{C_i\}_{i=1}^{L}
\]
of $K=12$ student audio tokens. If the final chunk contains fewer than
$K$ tokens, it is merged into the preceding chunk.

Let $e_i$ be the final student audio-token index of chunk $C_i$. The
student token-to-waveform decoder converts the cumulative token prefix
into waveform speech:
\[
\hat{r}_{\leq i}=D(a_1,\ldots,a_{e_i}).
\]
Cumulative-prefix decoding preserves the complete preceding acoustic
context, which would be unavailable if each chunk were decoded in
isolation. For the general scoring path analyzed in
Section~\ref{app:sec:computational_cost}, each cumulative waveform is
independently retokenized and evaluated by the teacher. The teacher
passes operate on complete cumulative prefixes rather than incrementally
extending a teacher acoustic key--value (KV) cache. Offline style
instructions are cached and reused; this conditioning reuse is distinct
from waveform-decoder state reuse and does not remove the per-prefix
teacher evaluations.

The TTS tokenizer converts the cumulative waveform prefix into a
teacher-side acoustic target sequence:
\[
\tilde{a}^{(i)}
=
E_T(\hat{r}_{\leq i})
=
\left(
\tilde{a}^{(i)}_1,\ldots,\tilde{a}^{(i)}_{M_i}
\right),
\]
where $M_i$ is the number of teacher-side acoustic tokens produced for
the $i$-th cumulative prefix. Since the student and teacher use
different acoustic tokenizers and token rates, their token boundaries
are not assumed to be aligned.

For $i\geq 2$, we define the teacher-token region newly introduced at
the current student chunk boundary as
\[
A_i=\{M_{i-1}+1,\ldots,M_i\}.
\]
The acoustic score assigned to $C_i$ is the mean teacher-forcing
negative log-likelihood over this newly introduced suffix:
\[
s_i
=
-\frac{1}{|A_i|}
\sum_{u\in A_i}
\log p_T
\left(
\tilde{a}^{(i)}_u
\mid
\tilde{a}^{(i)}_{<u},
r^{p},
z,
\hat{u}
\right).
\]
Earlier teacher-side tokens remain part of the teacher-forcing context
but do not contribute again to the score assigned to the current
student chunk.

\subsection{First-Chunk Exclusion}
\label{app:sec:first_chunk_exclusion}

We exclude the first student audio chunk $C_1$ from reward calibration
and from the TAPD optimization objective. The waveform associated with
the first chunk is typically very short, and its teacher-forcing NLL is
often substantially higher than that of later chunks. In the
heterogeneous student--teacher setting, this initial score is
particularly sensitive to startup tokens, tokenizer-rate mismatch, and
short-prefix boundary effects. It therefore provides a less reliable
estimate of paralinguistic compatibility.

The first cumulative prefix is nevertheless decoded and tokenized to
obtain $M_1$, which defines the teacher-side starting boundary of
$A_2$. Moreover, the waveform corresponding to $C_1$ remains part of
the cumulative acoustic context used to score every subsequent chunk.
Thus, first-chunk exclusion removes only its direct optimization signal
and does not remove its contextual contribution.

\subsection{Robust Teacher-Score Winsorization}
\label{app:sec:teacher_score_winsorization}

The token-level NLLs produced by the TTS teacher may occasionally
contain abnormally large spikes caused by acoustic-tokenization
mismatches, imperfect chunk boundaries, or local waveform-decoding
artifacts. Although the teacher-token NLLs are averaged within each
tail region to obtain a chunk score, a sufficiently large local spike
can still produce an unusually high chunk-level NLL. Because lower NLL
indicates stronger acoustic compatibility, such outliers would yield
disproportionately large negative preference signals and could distort
the reward-normalization statistics.

To reduce the influence of these outliers, we apply upper-tail
Winsorization independently within each utterance. Winsorization is
performed after first-chunk exclusion and before reward normalization.
Let
\[
\mathcal{S}_u
=
\{s_{u,1},s_{u,2},\ldots,s_{u,n_u}\}
\]
denote the valid post-skip chunk-level NLL scores of utterance \(u\),
where \(n_u\) is the number of retained chunks. Let
\(q_{\tau}(\mathcal{S}_u)\) denote the empirical
\(\tau\)-quantile of these scores. Each score is transformed as
\begin{equation}
    \widetilde{s}_{u,j}
    =
    \min
    \left(
        s_{u,j},
        q_{\tau}(\mathcal{S}_u)
    \right).
    \label{app:eq:teacher_score_winsorization}
\end{equation}

We use \(\tau=0.9\) in all experiments. Thus, scores above the
utterance-specific 90th-percentile threshold are replaced by that
threshold, while all remaining scores are left unchanged. Unlike
sample rejection, this operation retains every valid chunk and
preserves its training signal, while preventing a small number of
extreme teacher-score values from dominating reward calibration and
the resulting TAPD update.

The Winsorized scores \(\widetilde{s}_{u,j}\) are used consistently for
first-step within-utterance normalization, global EMA initialization,
subsequent EMA updates, and TAPD chunk-weight computation. For
notational simplicity, the following sections write \(s_{u,j}\) to
refer to the Winsorized score unless otherwise stated. Winsorization
affects only reward construction; it does not modify the TTS teacher
forward pass, the student rollout, the decoded waveform, or the
auxiliary SFT objective.

\subsection{Global EMA Reward Calibration}
\label{app:sec:tapd_reward_calibration}

Raw teacher NLL values may vary across utterances and over the course
of training. We therefore calibrate valid post-skip chunk scores using
global exponential moving average statistics. Let $\mu_t$ and
$\sigma_t^2$ denote the pre-update running mean and variance at
training step $t$. The normalized preference value for chunk $C_i$ is
\[
\tilde{\Delta}_i
=
\operatorname{clip}
\left(
\frac{\mu_t-s_i}{\sqrt{\max(\sigma_t^2,\epsilon)}+\epsilon},
-c,
c
\right),
\]
where $c=1.0$ and $\epsilon$ is a small numerical-stability constant.
Because lower NLL indicates greater compatibility, a chunk with
$s_i<\mu_t$ receives a positive preference value, whereas a chunk with
$s_i>\mu_t$ receives a negative value.

We then apply a symmetric dead zone:
\[
w_i
=
\tilde{\Delta}_i
\mathbb{I}
\left[
|\tilde{\Delta}_i|\geq d
\right],
\qquad d=0.5.
\]
This operation removes low-magnitude updates whose direction may be
dominated by teacher-score noise. The current batch is normalized using
the pre-update EMA statistics. Only after the optimization step are the
running statistics updated using the global moments of valid post-skip
chunk scores. We use $\beta=0.95$ and the mean and variance updates in
Equations~\ref{app:eq:ema_mean_update} and~\ref{app:eq:ema_variance_update}.
The variance update includes the mean-shift correction; initialization
and the complete update procedure are specified in
Section~\ref{app:sec:ema_initialization}.

\subsection{EMA Initialization and Cross-Step Updating}
\label{app:sec:ema_initialization}

TAPD employs a global chunk-level exponential moving average (EMA) to
normalize teacher scores across utterances and optimization steps. Let
\[
\mathcal{S}_{u}
=
\{s_{u,1},\ldots,s_{u,n_u}\}
\]
denote the valid chunk-level NLL scores of utterance \(u\), after
upper-tail score winsorization and first-chunk exclusion. Here, \(n_u\)
is the number of valid chunks retained for utterance \(u\). A lower NLL
indicates stronger agreement between the student-generated speech and
the conditioning information provided to the TTS teacher.

\paragraph{First-step normalization.}
At the beginning of training, historical global EMA statistics are not
yet available. We therefore use within-utterance normalization only for
the first optimization step. For each utterance \(u\), we compute
\begin{equation}
    \mu_u
    =
    \frac{1}{n_u}
    \sum_{j=1}^{n_u}
    s_{u,j},
    \qquad
    \sigma_u^2
    =
    \frac{1}{n_u}
    \sum_{j=1}^{n_u}
    \left(s_{u,j}-\mu_u\right)^2.
    \label{app:eq:first_step_utterance_moments}
\end{equation}
The initial uncalibrated chunk advantage is then
\begin{equation}
    A_{u,j}^{(1)}
    =
    \frac{
        \mu_u-s_{u,j}
    }{
        \sqrt{\max(\sigma_u^2,\epsilon)}+\epsilon
    },
    \label{app:eq:first_step_chunk_advantage}
\end{equation}
where \(\epsilon=10^{-6}\) is a numerical-stability constant. Consequently,
chunks with lower NLL than the corresponding utterance mean receive
positive advantages, whereas chunks with higher NLL receive negative
advantages. When only one valid chunk remains in an utterance, its
advantage is zero under this definition.

The same symmetric clipping and dead-zone operations used in later
steps are subsequently applied to \(A_{u,j}^{(1)}\):
\begin{equation}
    \widetilde{A}_{u,j}^{(1)}
    =
    \operatorname{clip}
    \left(
        A_{u,j}^{(1)},-c,c
    \right),
\end{equation}
\begin{equation}
    w_{u,j}^{(1)}
    =
    \widetilde{A}_{u,j}^{(1)}
    \mathbb{I}
    \left[
        \left|
        \widetilde{A}_{u,j}^{(1)}
        \right|
        \geq d
    \right].
\end{equation}

\paragraph{Direct initialization of global statistics.}
During the first optimization step, valid chunk scores are buffered
across all gradient-accumulation micro-steps. After the first model
update, the scores are aggregated across all distributed worker
processes. Let
\[
\mathcal{B}_1
=
\{s_i^{(1)}\}_{i=1}^{N_1}
\]
denote the resulting global collection of valid chunk scores. We
initialize the global EMA directly from their population mean and
variance:
\begin{equation}
    \mu_{\mathrm{EMA}}^{(1)}
    =
    \frac{1}{N_1}
    \sum_{i=1}^{N_1}
    s_i^{(1)},
    \label{app:eq:ema_mean_initialization}
\end{equation}
\begin{equation}
    v_{\mathrm{EMA}}^{(1)}
    =
    \frac{1}{N_1}
    \sum_{i=1}^{N_1}
    \left(
        s_i^{(1)}
        -
        \mu_{\mathrm{EMA}}^{(1)}
    \right)^2.
    \label{app:eq:ema_variance_initialization}
\end{equation}
The initial variance is lower-bounded by $\epsilon=10^{-6}$ before use:
$v_{\mathrm{EMA}}^{(1)}\leftarrow\max(v_{\mathrm{EMA}}^{(1)},\epsilon)$.
Direct initialization avoids the transient bias that would result from
gradually moving the running statistics away from arbitrary
zero-valued initial estimates.

\paragraph{Normalization after initialization.}
For every optimization step \(t>1\), each current chunk is normalized
using the EMA statistics accumulated before that step:
\begin{equation}
    A_i^{(t)}
    =
    \frac{
        \mu_{\mathrm{EMA}}^{(t-1)}
        -
        s_i^{(t)}
    }{
        \sqrt{\max\!\left(v_{\mathrm{EMA}}^{(t-1)},\epsilon\right)}
        +\epsilon
    }.
    \label{app:eq:ema_chunk_advantage}
\end{equation}
A chunk whose NLL is lower than the historical global mean therefore
receives a positive advantage, whereas a chunk with a higher NLL
receives a negative advantage. The calibrated optimization weight is
computed as
\begin{equation}
    \widetilde{A}_i^{(t)}
    =
    \operatorname{clip}
    \left(
        A_i^{(t)},-c,c
    \right),
\end{equation}
\begin{equation}
    w_i^{(t)}
    =
    \widetilde{A}_i^{(t)}
    \mathbb{I}
    \left[
        \left|
        \widetilde{A}_i^{(t)}
        \right|
        \geq d
    \right],
    \label{app:eq:ema_calibrated_weight}
\end{equation}
where \(c=1.0\) and \(d=0.5\) in the main experiments.

\paragraph{EMA update.}
Let
\[
\mathcal{B}_t
=
\{s_i^{(t)}\}_{i=1}^{N_t}
\]
be the valid chunk scores aggregated across devices and
gradient-accumulation micro-steps at optimization step \(t\). Their
current-step global mean and variance are
\begin{equation}
    \mu_{\mathcal{B}}^{(t)}
    =
    \frac{1}{N_t}
    \sum_{i=1}^{N_t}
    s_i^{(t)},
    \label{app:eq:batch_score_mean}
\end{equation}
\begin{equation}
    v_{\mathcal{B}}^{(t)}
    =
    \frac{1}{N_t}
    \sum_{i=1}^{N_t}
    \left(
        s_i^{(t)}
        -
        \mu_{\mathcal{B}}^{(t)}
    \right)^2.
    \label{app:eq:batch_score_variance}
\end{equation}

After the model update, the EMA mean is updated as
\begin{equation}
    \mu_{\mathrm{EMA}}^{(t)}
    =
    \beta
    \mu_{\mathrm{EMA}}^{(t-1)}
    +
    (1-\beta)
    \mu_{\mathcal{B}}^{(t)}.
    \label{app:eq:ema_mean_update}
\end{equation}
The corresponding variance update is
\begin{align}
    v_{\mathrm{EMA}}^{(t)}
    ={}&
    \beta
    v_{\mathrm{EMA}}^{(t-1)}
    +
    (1-\beta)
    v_{\mathcal{B}}^{(t)}
    \nonumber\\
    &+
    \beta(1-\beta)
    \left(
        \mu_{\mathcal{B}}^{(t)}
        -
        \mu_{\mathrm{EMA}}^{(t-1)}
    \right)^2.
    \label{app:eq:ema_variance_update}
\end{align}
The final term accounts for the difference between the historical and
current score means; omitting it would underestimate the variance when
the teacher-score distribution shifts during training. After every
update, we apply the numerical floor
\[
 v_{\mathrm{EMA}}^{(t)}\leftarrow
 \max\!\left(v_{\mathrm{EMA}}^{(t)},\epsilon\right),
 \qquad \epsilon=10^{-6}.
\]
The running standard deviation is
$\sigma_t=\sqrt{\max(v_{\mathrm{EMA}}^{(t)},\epsilon)}$;
normalized advantages use $\sigma_t+\epsilon$ in the denominator.

Importantly, the EMA statistics are updated only after all rewards for
the current optimization step have been computed. Thus, a chunk cannot
directly alter the normalization baseline used to determine its own
training weight. We use an EMA decay of \(\beta=0.95\), which smooths
short-term batch fluctuations while allowing the calibration reference
to adapt as the student policy changes.

\subsection{Entropy-Based Selective Distillation}
\label{app:sec:tapd_selective_distillation}

The teacher supplies one acoustic compatibility value for an entire
student chunk rather than exact supervision for each student audio
token. Applying this coarse-grained signal densely to every position
may overwrite reliable token decisions. TAPD therefore applies the
chunk feedback only to uncertain student audio-token positions.

For each valid assistant audio-token position $t$, we restrict the
student next-token distribution to the audio vocabulary
$\mathcal{V}_{\mathrm{audio}}$ and renormalize it:
\[
\bar{p}_{\theta}
(v\mid y_{<t},x)
=
\frac{
p_{\theta}(v\mid y_{<t},x)
}{
\sum_{v'\in\mathcal{V}_{\mathrm{audio}}}
p_{\theta}(v'\mid y_{<t},x)
},
\qquad
v\in\mathcal{V}_{\mathrm{audio}}.
\]
The corresponding audio-vocabulary entropy is
\[
H_t
=
-\sum_{v\in\mathcal{V}_{\mathrm{audio}}}
\bar{p}_{\theta}(v\mid y_{<t},x)
\log
\bar{p}_{\theta}(v\mid y_{<t},x).
\]

Within each completed rollout, we rank all valid assistant audio-token
positions according to $H_t$ and retain the highest-entropy
$\rho=0.5$ fraction. The selected set is treated as fixed during
backpropagation. For chunk $C_i$, the selective token-level loss is
\[
\ell_i^{\mathrm{sel}}
=
-\frac{1}{|P_i\cap S|}
\sum_{t\in P_i\cap S}
\log
\pi_{\theta}(y_t\mid y_{<t},x),
\]
where $P_i$ denotes the student-sequence positions belonging to chunk
$C_i$, and $S$ is the selected high-entropy position set. Chunks with
no selected positions are excluded from the TAPD loss. Chunks with
selected positions and zero calibrated weight remain in the averaging
denominator, although their weighted loss contribution is zero.

\subsection{Auxiliary Controllable-TTS SFT Supervision}
\label{app:sec:tapd_auxiliary_sft}

TAPD uses the same controllable-TTS synthetic supervision as the
Controllable-TTS SFT baseline. Starting from each original
query--response pair in the training set, we synthesize the
ground-truth assistant response using the controllable TTS teacher
corresponding to the current experimental setting. Specifically,
CosyVoice2-generated targets are used in the CosyVoice2 setting, while
CosyVoice3-generated targets are used in the CosyVoice3 setting.

For each training example, the TTS model is conditioned on the
ground-truth response text, the reference prompt speech, and the cached
style instruction \(z\). Importantly, this style instruction is exactly
the same instruction used to condition the TTS teacher during TAPD
on-policy scoring for the corresponding example. No separate style
parser, prompt template, or instruction re-inference is used for the
auxiliary SFT branch.

The resulting synthesized speech is converted into the student
acoustic-token representation and combined with the corresponding
assistant text to form the complete supervised target \(y^{\star}\).

The auxiliary SFT loss is computed by teacher forcing over the complete
reference assistant response:
\begin{equation}
    \mathcal{L}_{\mathrm{SFT}}
    =
    -\frac{1}{|\mathcal{T}_{\mathrm{SFT}}|}
    \sum_{t\in\mathcal{T}_{\mathrm{SFT}}}
    \log
    \pi_{\theta}
    \left(
        y^{\star}_t
        \mid
        x,y^{\star}_{<t}
    \right),
    \label{app:eq:tapd_auxiliary_sft}
\end{equation}
where $\mathcal{T}_{\mathrm{SFT}}$ contains the valid supervised
assistant positions, including assistant text, acoustic, and output
special-token positions. Prompt positions, padding positions, and
masked positions are excluded.

Importantly, the auxiliary SFT target is constructed from the original
ground-truth response and is independent of the student-generated
rollout used for TAPD reward computation. TAPD and the
Controllable-TTS SFT baseline therefore receive identical offline
synthetic supervision; their difference is that TAPD additionally uses
localized on-policy acoustic feedback from the same frozen TTS teacher.

\subsection{Combined Training Objective}
\label{app:sec:tapd_combined_objective}

Let $\mathcal{I}$ denote the set of non-initial chunks satisfying
$|P_i\cap S|>0$, including those with $w_i=0$. The TAPD objective is
\[
\mathcal{L}_{\mathrm{TAPD}}
=
\frac{1}{|\mathcal{I}|}
\sum_{i\in\mathcal{I}}
\operatorname{sg}(w_i)
\ell_i^{\mathrm{sel}},
\]
where $\operatorname{sg}(\cdot)$ denotes stop-gradient. If
$\mathcal{I}$ is empty, we set
$\mathcal{L}_{\mathrm{TAPD}}=0$.
Zero-weight chunks therefore remain in $|\mathcal{I}|$. Holding other
factors fixed, the effective acoustic-distillation contribution decreases
as fewer chunks receive nonzero weights. Excluding zero-weight chunks
from the denominator would instead rescale the remaining contributions
upward. This definition matches the normalization used in the main
experiments and does not change the reported results.

Following the notation in the main paper, we express the relative
training objective as
\[
\mathcal{L}_{\mathrm{relative}}
=
\mathcal{L}_{\mathrm{TAPD}}
+
\lambda_{\mathrm{SFT}}
\mathcal{L}_{\mathrm{SFT}},
\]
where the SFT loss is computed over the complete reference assistant
response. We use
\[
\lambda_{\mathrm{SFT}}=2.0.
\]
In the implementation, the complete relative objective is multiplied
by an overall loss scale of $0.5$. The loss passed to the optimizer is
therefore
\[
\mathcal{L}_{\mathrm{train}}
=
0.5
\left(
\mathcal{L}_{\mathrm{TAPD}}
+
2.0\mathcal{L}_{\mathrm{SFT}}
\right)
=
0.5\mathcal{L}_{\mathrm{TAPD}}
+
\mathcal{L}_{\mathrm{SFT}}.
\]

The sampled rollout, teacher-derived chunk weights, and entropy-based
selection set are treated as fixed for the current optimization step.
The auxiliary SFT term stabilizes training and preserves the student's
general response-generation behavior, while TAPD modifies uncertain
audio-token decisions using localized on-policy acoustic feedback.

\section{Baseline Implementations}
\label{app:sec:baseline_implementations}

This section describes the implementation of the trainable baselines.
All trainable methods use the shared optimization configuration in
Section~\ref{app:sec:experimental_setup}. For each TTS-teacher setting, the
corresponding controllable TTS model is used both to construct the
synthetic SFT targets and, when applicable, to provide teacher-score
feedback. Thus, comparisons within each teacher setting use the same
student initialization, training examples, synthetic SFT supervision,
optimizer configuration, and random seed.

\subsection{Shared Controllable-TTS SFT Supervision}
\label{app:sec:shared_sft_supervision}

All trainable baselines include the same supervised objective used by
TAPD. Starting from the 10,000 original query--response pairs, we use
the corresponding frozen TTS teacher to synthesize the ground-truth
assistant response. The TTS model is conditioned on the ground-truth
response text, the reference prompt speech, and the cached style
instruction produced during offline preprocessing. CosyVoice2 is used
to construct the synthetic targets in the CosyVoice2 setting, while
CosyVoice3 is used in the CosyVoice3 setting.

The resulting training target contains the complete assistant response,
including its valid text tokens, synthesized acoustic tokens, and
output special tokens. Prompt tokens are excluded from the supervised
loss. Let \(y^{\star}\) denote this synthetic target response. The
shared SFT objective is
\begin{equation}
    \mathcal{L}_{\mathrm{SFT}}
    =
    -\frac{1}{|\mathcal{T}_{\mathrm{SFT}}|}
    \sum_{t\in\mathcal{T}_{\mathrm{SFT}}}
    \log
    \pi_{\theta}
    \left(
        y^{\star}_t
        \mid
        x,y^{\star}_{<t}
    \right),
    \label{app:eq:shared_baseline_sft}
\end{equation}
where \(\mathcal{T}_{\mathrm{SFT}}\) contains all supervised assistant
positions. Unless otherwise stated, the implemented coefficient of
this auxiliary objective is \(1.0\) for every trainable method.

The \emph{Original Student} and \emph{TTS Teacher} results reported in
the main paper are evaluation-only references and do not involve
additional training.

\subsection{Controllable-TTS SFT}
\label{app:sec:controllable_tts_sft}

The Controllable-TTS SFT baseline removes all on-policy teacher-score
feedback and optimizes only the shared synthetic SFT objective:
\begin{equation}
    \mathcal{L}_{\mathrm{TTS\text{-}SFT}}
    =
    \mathcal{L}_{\mathrm{SFT}}.
    \label{app:eq:controllable_tts_sft}
\end{equation}
This baseline uses the same style-instruction inference pipeline,
reference prompt speech, and controllable TTS model as the
corresponding TAPD experiment. It therefore measures the improvement
obtained from on-policy acoustic feedback beyond conventional offline
imitation of controllable-TTS-generated speech.

\subsection{Sentence-Level TTS-NLL}
\label{app:sec:sentence_level_baseline}

We construct a sentence-level baseline to evaluate whether localized
chunk-level credit assignment is necessary. Given a student rollout,
we decode its complete valid student acoustic-token sequence into a
waveform. The frozen TTS teacher then evaluates this waveform once
under teacher forcing, conditioned on the complete rollout text, the
reference prompt speech, and the cached style instruction.

Let \(\ell_{u,k}^{T}\) denote the teacher NLL at its \(k\)-th valid
acoustic-token position for utterance \(u\). We define the
sentence-level score as
\begin{equation}
    S_u
    =
    \frac{1}{K_u}
    \sum_{k=1}^{K_u}
    \ell_{u,k}^{T},
    \label{app:eq:sentence_level_score}
\end{equation}
where \(K_u\) is the number of valid teacher acoustic tokens. Lower
\(S_u\) indicates stronger agreement with the rollout text, target
voice, and requested speaking style.

Because only one score is available for each rollout,
within-utterance normalization is not applicable. Sentence-level
scores are instead normalized using historical global EMA statistics:
\begin{equation}
    A_u^{(t)}
    =
    \frac{
        \mu_{\mathrm{EMA}}^{(t-1)}
        -
        S_u^{(t)}
    }{
        \sqrt{\max\!\left(v_{\mathrm{EMA}}^{(t-1)},\epsilon\right)}
        +\epsilon
    }.
    \label{app:eq:sentence_level_advantage}
\end{equation}
Thus, an utterance with lower NLL than the historical global average
receives a positive advantage, whereas an utterance with higher NLL
receives a negative advantage.

We apply the same symmetric clipping and dead-zone thresholds as in
TAPD:
\begin{equation}
    \widetilde{A}_u^{(t)}
    =
    \operatorname{clip}
    \left(
        A_u^{(t)},-c,c
    \right),
\end{equation}
\begin{equation}
    w_u^{(t)}
    =
    \widetilde{A}_u^{(t)}
    \mathbb{I}
    \left[
        \left|
        \widetilde{A}_u^{(t)}
        \right|
        \geq d
    \right],
    \label{app:eq:sentence_level_weight}
\end{equation}
where \(c=1.0\) and \(d=0.5\).

Before sentence-level EMA initialization, the fallback mean and standard
deviation are computed from the current utterance-level score. With one
score per utterance, the resulting normalized advantage is zero; the
initial acoustic-distillation weight is therefore zero and only the
shared SFT term contributes. After the first valid optimization step,
we initialize the EMA mean and population variance directly from all
valid utterance scores collected across gradient-accumulation micro-steps
and distributed workers. The variance is lower-bounded by
$\epsilon=10^{-6}$. On subsequent steps, rewards use the pre-update EMA
statistics, and the running statistics are updated after optimization
using Equations~\ref{app:eq:ema_mean_update}
and~\ref{app:eq:ema_variance_update}, with one score per utterance in place
of one score per valid chunk.

The scalar weight \(w_u\) is shared by all valid assistant
audio-token positions in the rollout. Let
\(\mathcal{A}_u\) denote these positions. The sentence-level acoustic
distillation loss is
\begin{equation}
    \mathcal{L}_{\mathrm{Sent}}
    =
    -\frac{1}{|\mathcal{U}|}
    \sum_{u\in\mathcal{U}}
    \frac{w_u}{|\mathcal{A}_u|}
    \sum_{t\in\mathcal{A}_u}
    \log
    \pi_{\theta}
    \left(
        y_{u,t}
        \mid
        x_u,y_{u,<t}
    \right),
    \label{app:eq:sentence_level_loss}
\end{equation}
where \(\mathcal{U}\) contains rollouts with nonzero weights and at
least one valid audio-token position. If \(\mathcal{U}\) is empty,
we set \(\mathcal{L}_{\mathrm{Sent}}=0\).

To match the implemented reward-to-SFT scaling used by TAPD, the final
training objective is
\begin{equation}
    \mathcal{L}
    =
    0.5\mathcal{L}_{\mathrm{Sent}}
    +
    \mathcal{L}_{\mathrm{SFT}}.
    \label{app:eq:sentence_level_total_loss}
\end{equation}
No token-level entropy filtering is applied. Consequently, this
baseline preserves the same TTS-derived supervision and auxiliary SFT
objective as TAPD, while replacing localized chunk feedback with a
single sentence-level weight shared by the entire acoustic trajectory.

\subsubsection{Sentence-Level Control with Matched Token Selection and Optimization}
\label{app:sec:matched_sentence_control}

To separate the effect of feedback granularity from token selection and
optimization choices, we additionally evaluate a sentence-level control
under the CosyVoice3 teacher setting. This control matches TAPD's 50\%
entropy-based audio-token selection, EMA calibration, clipping and
dead-zone rules, loss aggregation, and acoustic-distillation/SFT
coefficients. Unlike TAPD's localized chunk weights, it uses a single
sentence-level acoustic weight throughout each rollout. This control
is distinct from the original Sentence-Level TTS-NLL baseline in
Table~\ref{tab:main_results}, which does not use entropy filtering.

\begin{table}[tp]
\centering
\small
\setlength{\tabcolsep}{4pt}
\caption{Feedback-granularity comparison with matched token selection
and optimization under the CosyVoice3 teacher. Averages are unweighted
across the four evaluation subsets.}
\label{app:tab:matched_sentence_control}
\begin{tabular}{lcc}
\toprule
\textbf{Method} & \textbf{MOS Avg.} $\uparrow$ & \textbf{IFR Avg.} $\uparrow$ \\
\midrule
Matched sentence-level control & 4.141 & 64.06 \\
TAPD (chunk-level feedback) & 4.235 & 68.90 \\
\bottomrule
\end{tabular}
\end{table}

As shown in Table~\ref{app:tab:matched_sentence_control}, TAPD retains a
4.84-percentage-point average IFR advantage over the matched
sentence-level control. This comparison supports the benefit of
localized acoustic feedback beyond entropy-based token selection alone.

\subsection{TTS-GRPO}
\label{app:sec:tts_grpo_baseline}

We additionally construct a TTS-guided Group Relative Policy
Optimization baseline to compare TAPD with a conventional group-based
policy-gradient objective. For each input prompt \(x\), the current
student policy samples a group of \(G\) independent responses:
\begin{equation}
    \mathcal{Y}
    =
    \{y_1,\ldots,y_G\},
    \qquad
    G=3.
\end{equation}
We use three rollouts per prompt to provide a group-relative comparison
while limiting the substantially higher generation and TTS-scoring
cost of group-based optimization. The rollout sampling
settings are specified in
Section~\ref{app:sec:rollout_generation_settings}.

Each sampled acoustic-token sequence is decoded into a complete
waveform and evaluated once by the frozen TTS teacher. As in the
sentence-level baseline, teacher scoring is conditioned on the
complete rollout text, reference prompt speech, and cached style
instruction. Let \(\ell_{g,k}^{T}\) denote the teacher NLL of the
\(k\)-th valid teacher acoustic token for rollout \(g\). Its
sentence-level score is
\begin{equation}
    S_g
    =
    \frac{1}{K_g}
    \sum_{k=1}^{K_g}
    \ell_{g,k}^{T},
    \label{app:eq:grpo_sentence_nll}
\end{equation}
where \(K_g\) is the number of valid teacher acoustic tokens. Since
lower NLL indicates stronger acoustic compatibility, we define
\begin{equation}
    R_g=-S_g.
\end{equation}

Following GRPO, rewards are standardized within each rollout group
without a learned value model. We compute
\begin{equation}
    \overline{S}
    =
    \frac{1}{G}
    \sum_{g=1}^{G}S_g,
\end{equation}
\begin{equation}
    \sigma_S
    =
    \sqrt{
        \frac{1}{G}
        \sum_{g=1}^{G}
        \left(
            S_g-\overline{S}
        \right)^2
    },
\end{equation}
and define the group-relative advantage as
\begin{equation}
    A_g
    =
    \frac{
        \overline{S}-S_g
    }{
        \sigma_S+\epsilon
    }.
    \label{app:eq:grpo_group_advantage}
\end{equation}
Thus, rollouts with lower NLL than the group average receive positive
advantages, while those with higher NLL receive negative advantages.
If all group scores are identical, we set all group advantages to zero.

Let \(\widetilde{\pi}_{\theta}\) denote the policy distribution after
applying the same transformations used during rollout sampling,
including temperature scaling, top-\(k\) truncation, nucleus
truncation, repetition penalty, and audio-vocabulary masking. For each
valid assistant audio-token position \(t\), the importance ratio is
\begin{align}
    r_{g,t}(\theta)
    =
    \exp\Big(
        &\log
        \widetilde{\pi}_{\theta}
        (y_{g,t}\mid x,y_{g,<t})
        \nonumber\\
        &-
        \log
        \widetilde{\pi}_{\mathrm{old}}
        (y_{g,t}\mid x,y_{g,<t})
    \Big).
    \label{app:eq:grpo_ratio}
\end{align}
The old-policy log-probabilities are recomputed without gradients using
the policy snapshot employed to generate the rollout. The current and
old log-probabilities are evaluated under the same transformed
distribution as the rollout-generation process.

Let \(\mathcal{A}_g\) denote the valid assistant audio-token positions
of rollout \(g\). Text tokens, prompt tokens, masked positions, padded
positions, and post-EOS positions are excluded. The unscaled clipped
GRPO objective is
\begin{equation}
    \bar{r}_{g,t}(\theta)
    =
    \operatorname{clip}
    \left(
        r_{g,t}(\theta),
        1-\varepsilon,
        1+\varepsilon
    \right).
    \label{app:eq:grpo_clipped_ratio}
\end{equation}

\begin{equation}
\begin{aligned}
    \widehat{\mathcal{L}}_{\mathrm{GRPO}}
    ={}&
    -\frac{1}{G}
    \sum_{g=1}^{G}
    \frac{1}{|\mathcal{A}_g|}
    \sum_{t\in\mathcal{A}_g}
    \min\Bigl(
        r_{g,t}(\theta)A_g,
        \bar{r}_{g,t}(\theta)A_g
    \Bigr).
\end{aligned}
\label{app:eq:tts_grpo_unscaled_loss}
\end{equation}
where $\bar{r}_{g,t}(\theta)$ is the importance ratio clipped to
$[1-\varepsilon,1+\varepsilon]$, and we set
$\varepsilon=0.2$.

We scale the GRPO objective by
\(\alpha_{\mathrm{GRPO}}=0.1\):
\begin{equation}
    \mathcal{L}_{\mathrm{GRPO}}
    =
    0.1
    \widehat{\mathcal{L}}_{\mathrm{GRPO}}.
    \label{app:eq:tts_grpo_scaled_loss}
\end{equation}
This relatively conservative scale was selected because larger GRPO
weights produced overly aggressive policy updates and degraded
performance in development experiments.

The final TTS-GRPO training objective is
\begin{equation}
    \mathcal{L}
    =
    \mathcal{L}_{\mathrm{GRPO}}
    +
    \mathcal{L}_{\mathrm{SFT}}.
    \label{app:eq:tts_grpo_total_loss}
\end{equation}
The shared SFT loss is computed from the same controllable-TTS synthetic
targets used by TAPD, Sentence-Level TTS-NLL, and Controllable-TTS SFT.
We disable the reference-model KL regularizer
(\(\beta_{\mathrm{KL}}=0\)) and do not apply token-level entropy
selection. Therefore, every valid assistant audio token in a rollout
shares the same sentence-level group advantage.

\subsection{Summary of Baseline-Specific Settings}
\label{app:sec:baseline_setting_summary}

Table~\ref{app:tab:baseline_specific_settings} summarizes the
method-specific configurations of the trainable baselines. The
baselines are designed to isolate two factors that distinguish TAPD
from conventional alternatives: the temporal granularity of
TTS-derived feedback and the optimization objective used to incorporate
that feedback. Controllable-TTS SFT represents purely offline
supervision, Sentence-Level TTS-NLL retains on-policy teacher feedback
but collapses it into a single utterance-level score, and TTS-GRPO
applies the same type of utterance-level TTS score through the
group-relative policy-optimization framework introduced by
Shao et al.~\cite{DBLP:journals/corr/abs-2402-03300}.

To ensure a controlled comparison, all trainable methods use the same
student initialization, training examples, controllable-TTS-generated
SFT targets, auxiliary SFT coefficient, optimizer configuration, and
random seed. Method-specific differences are therefore restricted to
the form and granularity of the additional on-policy objective.

\begin{table*}[tp]
\centering
\small
\setlength{\tabcolsep}{4pt}
\renewcommand{\arraystretch}{1.12}
\caption{Method-specific configurations of the trainable baselines.
All methods use the same student initialization, 10,000-example
training set, controllable-TTS synthetic SFT supervision, and shared
optimization configuration.}
\label{app:tab:baseline_specific_settings}
\begin{tabularx}{\textwidth}{
    >{\raggedright\arraybackslash}p{0.19\textwidth}
    >{\raggedright\arraybackslash}p{0.18\textwidth}
    >{\raggedright\arraybackslash}p{0.17\textwidth}
    >{\raggedright\arraybackslash}X
}
\toprule
\textbf{Method}
& \textbf{Teacher feedback}
& \textbf{Objective scale}
& \textbf{Additional settings} \\
\midrule

Controllable-TTS SFT
& None
& SFT: \(1.0\)
& Complete controllable-TTS synthetic assistant response. \\

Sentence-Level TTS-NLL
& One full-utterance NLL
& Reward / SFT: \(0.5/1.0\)
& Global sentence-score EMA; \(c=1.0\); \(d=0.5\);
no entropy filtering. \\

TTS-GRPO
& Group-relative full-utterance NLL
& GRPO / SFT: \(0.1/1.0\)
& \(G=3\); ratio clip \(\varepsilon=0.2\);
\(\beta_{\mathrm{KL}}=0\); no entropy filtering. \\

\bottomrule
\end{tabularx}
\end{table*}

The resulting comparison forms a progression from offline imitation to
sentence-level on-policy supervision and group-relative policy
optimization. In contrast, TAPD preserves localized chunk-level credit
assignment and applies the resulting feedback selectively to uncertain
audio-token positions. Thus, the comparison separates the benefit of
TTS-derived feedback itself from the benefit of TAPD's localized and
selective optimization design.

\section{Inference Configuration and Decoding Robustness}
\label{app:sec:inference_and_decoding}

This section describes the inference configuration used to evaluate all
student checkpoints. We further explain why stochastic decoding is used
instead of greedy decoding and examine the sensitivity of the reported
results to the decoding seed. Unless otherwise stated, all student-side
methods, including the Original Student, Controllable-TTS SFT,
Sentence-Level TTS-NLL, TTS-GRPO, and TAPD, are evaluated using the same
inference configuration.

Training rollouts use the same sampling hyperparameters as evaluation,
with a shorter generation cap of 300 rather than 800 new tokens for
training efficiency. The training-time settings and rollout counts are
specified in Section~\ref{app:sec:rollout_generation_settings}.

\subsection{Evaluation-Time Inference Configuration}
\label{app:sec:evaluation_inference_configuration}

Table~\ref{app:tab:inference_configuration} summarizes the decoding
configuration used for the main evaluation. We use stochastic nucleus
sampling with a temperature of \(0.7\), a top-\(p\) threshold of
\(0.9\), top-\(k=50\) (the inherited \texttt{GenerationConfig} default),
and a repetition penalty of \(1.05\). The decoding seed is
fixed to \(42\), matching the default training seed, for all main
results.

\begin{table}[tp]
\centering
\small
\setlength{\tabcolsep}{4pt}
\renewcommand{\arraystretch}{1.10}
\caption{Evaluation-time inference configuration shared by all student
checkpoints.}
\label{app:tab:inference_configuration}
\begin{tabularx}{\columnwidth}{
    >{\raggedright\arraybackslash}p{0.45\columnwidth}
    >{\raggedright\arraybackslash}X
}
\toprule
\textbf{Configuration} & \textbf{Value} \\
\midrule
Maximum new tokens
& 800 \\

Maximum audio duration
& No separate limit (\texttt{0}) \\

Audio-token rate
& 25 tokens per second \\

Decoding strategy
& Stochastic sampling \\

Sampling temperature
& 0.7 \\

Nucleus threshold
& \(p=0.9\) \\

Top-\(k\) threshold
& 50 (default) \\

Repetition penalty
& 1.05 \\

Decoding seed
& 42 \\

Stopping rule
& \texttt{<|EOT|>} or the token cap \\

Skip existing outputs
& Enabled \\

\bottomrule
\end{tabularx}
\end{table}

The maximum generation length is 800 newly generated multimodal tokens.
We do not impose an additional audio-duration limit; audio generation
stops when the model emits \texttt{<|EOT|>} or reaches the 800-token
limit, whichever occurs first. The student generates audio at a nominal
rate of 25 acoustic tokens per second.

The \texttt{skip-existing} option is an execution-level setting that
prevents previously generated output files from being overwritten. It
does not alter the student distribution or the generated response.

\subsection{Rationale for Stochastic Decoding}
\label{app:sec:stochastic_decoding_rationale}

We use stochastic decoding because greedy decoding is unreliable for
the Step-Audio-2-mini student. In preliminary evaluation, greedy
decoding frequently entered degenerate acoustic trajectories,
qualitatively manifested as repetitive or noise-like audio segments.
Once such a high-probability but acoustically unstable trajectory is
selected, deterministic decoding cannot recover by exploring an
alternative continuation.

This behavior is particularly problematic for paralinguistic
generation. Multiple acoustic realizations can be semantically valid,
and the highest-probability token at every position does not
necessarily produce the most coherent utterance-level acoustic
trajectory. Controlled stochastic sampling provides limited
exploration among plausible continuations while the temperature,
nucleus threshold, and repetition penalty constrain excessively
unlikely or repetitive outputs.

We therefore use the same stochastic decoding configuration for every
student checkpoint. This ensures that differences among training
methods are not confounded by method-specific inference strategies.

\subsection{Comparison with Greedy Decoding}
\label{app:sec:greedy_decoding_comparison}

We empirically compare greedy decoding with the stochastic inference
configuration in Table~\ref{app:tab:greedy_vs_sampling}. Both evaluations
use the original Step-Audio-2-mini checkpoint without task-specific
post-training. To match the reporting convention of the main paper,
the average column is the unweighted mean over the emotion, speed,
volume, and composite subsets.

\begin{table*}[tp]
\centering
\scriptsize
\setlength{\tabcolsep}{3.2pt}
\renewcommand{\arraystretch}{1.10}
\caption{Comparison of greedy and stochastic decoding for the original
Step-Audio-2-mini student. Avg.\ denotes the unweighted average over
the four subsets.}
\label{app:tab:greedy_vs_sampling}
\begin{tabular}{
    l
    ccccc
    ccccc
}
\toprule
&
\multicolumn{5}{c}{\textbf{MOS \(\uparrow\)}} &
\multicolumn{5}{c}{\textbf{IFR (\%) \(\uparrow\)}} \\
\cmidrule(lr){2-6}
\cmidrule(lr){7-11}
\textbf{Decoding}
& \textbf{Emotion}
& \textbf{Speed}
& \textbf{Volume}
& \textbf{Composite}
& \textbf{Avg.}
& \textbf{Emotion}
& \textbf{Speed}
& \textbf{Volume}
& \textbf{Composite}
& \textbf{Avg.} \\
\midrule

Greedy
& 2.486
& 3.085
& 2.949
& 2.773
& 2.823
& 25.11
& 70.41
& 66.33
& 47.47
& 52.33 \\

Stochastic
& 3.024
& 3.990
& 3.878
& 3.780
& 3.668
& 29.71
& 82.67
& 70.33
& 57.00
& 59.93 \\

\bottomrule
\end{tabular}
\end{table*}

Stochastic decoding improves the unweighted average MOS from \(2.823\)
to \(3.668\), an absolute increase of \(0.845\). It also improves the
average IFR from \(52.33\%\) to \(59.93\%\), corresponding to a
\(7.60\)-percentage-point gain. The improvement is consistent across
all four subsets and is particularly pronounced for speaking rate and
composite-style instructions.

The greedy-decoding run contains 1,380 available generated responses,
whereas the stochastic-decoding run contains the complete set of 1,400
responses. All available outputs in both runs were successfully scored
by the automatic evaluator. Because the available generation counts
are not identical, this experiment is intended primarily as a
diagnostic analysis motivating the shared evaluation-time decoding
strategy rather than as a comparison between training methods.

\subsection{Sensitivity to Decoding Seeds}
\label{app:sec:decoding_seed_sensitivity}

Stochastic decoding introduces sampling variability even when the
model checkpoint and all decoding hyperparameters are fixed. To
quantify this effect, we evaluated the original Step-Audio-2-mini
checkpoint using five distinct decoding seeds while keeping the model,
test set, and inference configuration unchanged. Due to the cost of
API-based evaluation, this analysis was conducted only for the
original student rather than repeated for every post-trained
checkpoint.

Across the five decoding runs, the example-weighted overall MOS values
were \(3.4682\), \(3.4751\), \(3.4490\), \(3.4561\), and \(3.4629\).
The corresponding IFR values were \(51.71\%\), \(52.88\%\),
\(49.60\%\), \(51.33\%\), and \(50.96\%\). The resulting mean and
sample standard deviation were \(3.4623 \pm 0.0102\) for MOS and
\(51.30\% \pm 1.19\) percentage points for IFR. The MOS values ranged
from \(3.4490\) to \(3.4751\), while the IFR values ranged from
\(49.60\%\) to \(52.88\%\).

These results indicate that the selected stochastic decoding
configuration produces stable overall speech-quality scores.
Paralinguistic instruction following exhibits somewhat greater
sampling variability, but the variation remains moderate relative to
the performance differences observed among the main training methods.
We therefore use a fixed decoding seed of \(42\) for all main-paper
comparisons, ensuring that every student checkpoint is evaluated under
the same sampled decoding condition.

The aggregate scores reported in this seed analysis are the
example-weighted overall summaries produced by the evaluation pipeline.
In contrast, the main paper reports the unweighted mean over the
emotion, speed, volume, and composite subsets. The absolute aggregate
values from the two reporting conventions should therefore not be
compared directly.

\section{TTS Teacher-Score Validation}
\label{app:sec:tts_score_validation}

We conduct controlled analyses to examine whether teacher-forcing NLL
from the controllable TTS teacher provides a meaningful compatibility
signal for student-generated speech. We evaluate both sentence-level
instruction discrimination and localized acoustic perturbation
detection. In each analysis, the examples are evenly divided between
CosyVoice2 and CosyVoice3.

\subsection{Sentence-Level Instruction Discrimination}
\label{app:sec:sentence_level_score_validation}

We first evaluate whether sentence-level TTS NLL can distinguish
between matched and mismatched paralinguistic instructions. We randomly
sample 300 texts and construct a matched instruction and a conflicting
instruction for the corresponding speaking-rate, emotion, or volume
condition. The speech content and other conditioning information are
kept unchanged.

For each pair, the complete utterance is evaluated under both
instructions. A prediction is considered correct when the TTS teacher
assigns a lower sentence-level mean NLL to the matched instruction than
to the conflicting instruction. Table~\ref{app:tab:sentence_score_validation}
reports the resulting pairwise discrimination accuracy.

\begin{table}[tp]
\centering
\small
\setlength{\tabcolsep}{6pt}
\renewcommand{\arraystretch}{1.10}
\caption{Pairwise discrimination accuracy of sentence-level TTS NLL
under matched and conflicting paralinguistic instructions.}
\label{app:tab:sentence_score_validation}
\begin{tabular}{lc}
\toprule
\textbf{Instruction dimension}
& \textbf{Accuracy (\%)} \\
\midrule
Speaking rate & 61.33 \\
Emotion       & 68.33 \\
Volume        & 64.00 \\
\bottomrule
\end{tabular}
\end{table}

Sentence-level NLL performs consistently above chance across all three
dimensions, indicating that the teacher score contains information
about the requested paralinguistic condition. However, the accuracy
remains between \(61.33\%\) and \(68.33\%\), suggesting that averaging
NLL over the complete utterance can dilute localized acoustic
mismatches.

\subsection{Localized Perturbation Validation}
\label{app:sec:localized_score_validation}

We further examine whether the TTS teacher score is sensitive to
localized acoustic or lexical errors. We construct 150 paired
clean--perturbed examples, evenly divided between CosyVoice2 and
CosyVoice3. The examples cover three perturbation types, with 50
examples per type and 25 examples evaluated by each TTS teacher.

The first type locally increases or decreases the waveform amplitude.
Because localized emotion editing cannot be independently controlled
without introducing additional acoustic changes, we use local loudness
variation as a controllable proxy for emotion- and volume-related
intensity changes. The second type locally accelerates or slows a
selected speech segment while leaving the remainder of the utterance
unchanged. The third type replaces one spoken word with an incorrect
word while retaining the original conditioning text.

For every pair, the response text, reference speech, style instruction,
speaker identity, and unmodified waveform regions are kept unchanged.
We then compare the teacher score in the perturbed region with that of
the corresponding clean utterance. A perturbation is considered
successfully detected when the localized mean NLL, and equivalently its
perplexity, increases after modification.

The teacher score increases at the modified location in 143 out of 150
examples, corresponding to an overall detection accuracy of
\(95.33\%\). By perturbation type, it detects 46 out of 50 volume
perturbations (92\%), all 50 speaking-rate perturbations (100\%), and
47 out of 50 pronunciation perturbations (94\%).
Table~\ref{app:tab:localized_perturbation_breakdown} summarizes this breakdown.
These tests measure sensitivity to the specified local perturbations;
they do not directly measure chunk-level emotion discrimination.

\begin{table}[tp]
\centering
\small
\setlength{\tabcolsep}{5pt}
\caption{Localized perturbation detection by type, pooled across the
two TTS teachers (25 examples per type per teacher).}
\label{app:tab:localized_perturbation_breakdown}
\begin{tabular}{lcc}
\toprule
\textbf{Perturbation} & \textbf{Detected / Total} & \textbf{Accuracy} \\
\midrule
Volume & 46 / 50 & 92\% \\
Speaking rate & 50 / 50 & 100\% \\
Pronunciation & 47 / 50 & 94\% \\
\midrule
Overall & 143 / 150 & 95.33\% \\
\bottomrule
\end{tabular}
\end{table}

This result shows that teacher-forcing NLL is highly
sensitive to localized pronunciation, speaking-rate, and acoustic
intensity inconsistencies. In contrast to a single utterance-level
average, which can dilute a short erroneous region, localized scoring
preserves the temporal position of the mismatch. This observation
supports TAPD's use of cumulative-prefix scoring and tail-region
attribution for assigning feedback along the student's acoustic
trajectory.

\section{Chunk-Level Validation}
\label{app:sec:chunk_level_validation}

TAPD attributes the score of each cumulative speech prefix to the
newly introduced tail region in the TTS teacher's token space. Because
the student and TTS teacher use different acoustic tokenizations, this
construction requires the teacher-token length to remain monotonic as
the student prefix grows. We therefore examine both teacher-length
monotonicity and token-level stability across adjacent cumulative
prefixes.

\subsection{Teacher-Length Monotonicity}
\label{app:sec:teacher_length_monotonicity}

We randomly sample 100 utterances and segment their student acoustic
sequences using the default chunk size \(K=12\). This produces 2,035
adjacent cumulative-prefix comparisons. Let \(M_i\) denote the number
of teacher acoustic tokens obtained by tokenizing the cumulative
waveform up to student chunk \(i\).

As summarized in Table~\ref{app:tab:prefix_boundary_stability}, all 2,035
adjacent prefix pairs satisfy
\[
    M_i > M_{i-1}.
\]
No pair produces an unchanged or decreased teacher-token length.
Moreover, strict monotonicity holds for every sampled utterance.

\begin{table}[tp]
\centering
\small
\setlength{\tabcolsep}{5pt}
\renewcommand{\arraystretch}{1.10}
\caption{Teacher-token length stability across adjacent cumulative
prefixes with student chunk size \(K=12\).}
\label{app:tab:prefix_boundary_stability}
\begin{tabular}{lc}
\toprule
\textbf{Statistic} & \textbf{Result} \\
\midrule
Sampled utterances & 100 \\
Adjacent prefix pairs & 2,035 \\
Strict increase: \(M_i>M_{i-1}\) & 100.00\% \\
Unchanged length: \(M_i=M_{i-1}\) & 0.00\% \\
Length decrease: \(M_i<M_{i-1}\) & 0.00\% \\
Utterances monotonic throughout & 100.00\% \\
\bottomrule
\end{tabular}
\end{table}

Consequently, the tail region
\[
    \mathcal{A}_i
    =
    \{M_{i-1}+1,\ldots,M_i\}
\]
is non-empty for every observed chunk. This result supports the
practical validity of using consecutive teacher-token lengths to
define localized score regions under the default chunk configuration.

\subsection{Retokenization Stability}
\label{app:sec:prefix_retokenization_stability}

Although the cumulative waveform at step \(i\) contains the complete
waveform from step \(i-1\), the TTS tokenizer is applied independently
to each cumulative prefix. Appending a new speech segment may therefore
alter some token assignments in the previously observed region.

To quantify this effect, we compare the teacher-token sequence of each
old prefix with the corresponding token positions obtained after
retokenizing the extended prefix. Across all adjacent prefix pairs,
\(84.42\%\) of the compared token positions remain identical. Thus, the
majority of the previously generated teacher-token prefix is preserved,
although exact token-level prefix invariance does not always hold.

Importantly, these local token changes never cause the teacher-token
length to remain constant or decrease in our evaluation. Therefore,
while heterogeneous tokenization can modify individual tokens near or
within the previous prefix, it does not invalidate the length-based
tail construction used by TAPD. These observations motivate our use of
cumulative-prefix scoring: it does not require a fixed one-to-one
alignment between student and teacher acoustic tokens, while still
providing a consistently defined newly introduced teacher-token region
for localized feedback.

\section{Human Evaluation Validation}
\label{app:sec:human_evaluation_validation}

We conduct a human evaluation to validate the IFR labels produced by
the UltraVoice-based automatic evaluator. We focus on IFR rather than
MOS for two reasons. First, manual speech evaluation is costly, while
MOS requires annotators to distinguish among multiple ordinal quality
levels and generally requires more extensive calibration and repeated
ratings. Second, IFR directly measures whether the generated speech
follows the requested paralinguistic instruction and is therefore more
closely aligned with the primary objective of this work.

\subsection{Annotation Protocol}
\label{app:sec:human_annotation_protocol}

We randomly sample 100 previously evaluated utterances, consisting of
50 examples assigned \(\mathrm{IFR}=0\) and 50 examples assigned
\(\mathrm{IFR}=1\) by the automatic evaluator. The automatic labels are
hidden from the human annotators. Two annotators independently evaluate
all 100 examples, resulting in 200 human judgments in total.

For each example, the annotator listens to the generated speech and
determines whether it follows the associated paralinguistic instruction.
The human decision is represented using the same binary convention as
the automatic metric: \(\mathrm{IFR}=1\) indicates successful
instruction following, whereas \(\mathrm{IFR}=0\) indicates failure.

\subsection{Agreement with Automatic IFR Evaluation}
\label{app:sec:human_automatic_agreement}

For the 50 examples automatically labeled as \(\mathrm{IFR}=0\), the
two annotators agree with the automatic evaluator on \(96\%\) and
\(92\%\) of the examples, respectively. These values correspond to
48 and 46 matching judgments out of 50. For the 50 examples
automatically labeled as \(\mathrm{IFR}=1\), their agreement rates are
\(84\%\) and \(78\%\), corresponding to 42 and 39 matching judgments.

The first annotator therefore agrees with the automatic labels on
90 out of 100 examples, yielding an overall agreement rate of
\(90\%\). The second annotator agrees on 85 out of 100 examples,
yielding an agreement rate of \(85\%\). Aggregating both annotators
gives 175 matching judgments out of 200, corresponding to an overall
human--automatic agreement rate of \(87.5\%\).

When aggregated by automatic label, agreement is \(94\%\) for
\(\mathrm{IFR}=0\) and \(81\%\) for \(\mathrm{IFR}=1\). The lower
agreement for positive predictions suggests that determining whether
an instruction has been fully satisfied can be more subjective than
identifying an evident instruction-following failure. Nevertheless,
the overall agreement indicates that the automatic IFR evaluator is
broadly consistent with human judgments and is suitable for the
large-scale comparisons reported in this work.

\section{Computational Cost Analysis}
\label{app:sec:computational_cost}

Compared with standard SFT, the additional computation introduced by
on-policy training mainly consists of student rollout generation,
waveform reconstruction, and TTS-teacher evaluation. Since the
auxiliary SFT loss, backpropagation, and optimizer update are shared
across the trainable methods, we exclude these common operations and
compare only the method-specific cost of constructing on-policy
feedback.

We profile a representative response containing 300 generated tokens,
including 240 acoustic tokens. For TAPD, the default chunk size
\(K=12\) produces 20 cumulative prefixes. All measurements use the
same student model, waveform decoder, TTS teacher, hardware
environment, and response length.

\subsection{Method-Specific Feedback Cost}
\label{app:sec:method_specific_feedback_cost}

Sentence-Level TTS-NLL requires one student rollout, one full-response
waveform decoding pass, and one full-response teacher evaluation.
We use its feedback-construction latency as the normalized reference
cost.

TAPD also requires only one student rollout. In its general
tokenizer-independent implementation, however, each cumulative waveform
prefix is independently retokenized and evaluated by the TTS teacher.
These are complete prefix evaluations, not incremental extensions of a
teacher acoustic KV cache; the reported costs include waveform
reconstruction and teacher evaluation. Under the
default \(K=12\) configuration, the resulting feedback-construction
cost is \(2.76\times\) that of the sentence-level baseline.

TTS-GRPO requires \(G\) independently sampled student responses and a
full-response teacher evaluation for every response. Its cost
therefore increases approximately linearly with the group size. With
the relatively small group size \(G=3\) used in our main experiments,
its measured feedback-construction cost is \(3.00\times\) that of the
sentence-level baseline, which remains higher than the general TAPD
implementation.

\begin{table}[tp]
\centering
\small
\setlength{\tabcolsep}{4pt}
\renewcommand{\arraystretch}{1.10}
\caption{Normalized method-specific cost of constructing on-policy
feedback. Common SFT computation, backpropagation, and optimizer
updates are excluded.}
\label{app:tab:normalized_computational_cost}
\begin{tabularx}{\columnwidth}{
    >{\raggedright\arraybackslash}X
    >{\centering\arraybackslash}p{0.29\columnwidth}
}
\toprule
\textbf{Method} & \textbf{Normalized cost} \\
\midrule
Sentence-Level TTS-NLL
& \(1.00\times\) \\

TAPD, aligned fast path
& \(1.00\times\) \\

TAPD, general cumulative scoring
& \(2.76\times\) \\

TTS-GRPO, \(G=3\)
& \(3.00\times\) \\
\bottomrule
\end{tabularx}
\end{table}

This comparison is conservative for GRPO because the main experiments
already use a small group size of \(G=3\). Larger groups would increase
both rollout generation and teacher-evaluation cost approximately
linearly. In contrast, TAPD uses a single student trajectory, and its
cost is controlled by the number of chunks: increasing the chunk size
reduces the number of cumulative teacher evaluations, trading finer
credit assignment for greater computational efficiency.

\subsection{Efficient Scoring under Temporal Alignment}
\label{app:sec:aligned_fast_path}

The cumulative implementation is designed for the general case in
which the student and TTS teacher use different acoustic tokenizers.
It avoids assuming an exact one-to-one correspondence between their
token sequences and therefore supports heterogeneous speech-model
architectures.

When a reliable temporal mapping between student and teacher acoustic
representations is available, an aligned full-response variant can
reduce scoring cost. The complete student response is decoded once, and
a single teacher-forced full-response evaluation produces token-level
NLL values. Chunk scores are then obtained by aggregating these values
over the corresponding temporal regions.

This fast path is an alternative scoring implementation, not an exact
acceleration of cumulative-prefix tail scoring. Retokenizing a complete
response can produce different teacher acoustic targets from those
obtained from its cumulative prefixes, as discussed in
Section~\ref{app:sec:prefix_retokenization_stability}. Temporal alignment
therefore does not guarantee identical chunk scores or calibrated
weights. The shared nominal acoustic-token rate of 25~Hz facilitates
boundary mapping but does not establish scoring equivalence.

The aligned variant has approximately the same feedback-construction
cost as the sentence-level baseline in the latency comparison, but that
comparison does not establish equivalent training behavior or quality.
All main TAPD results and the end-to-end training times in
Section~\ref{app:sec:end_to_end_training_time} use cumulative-prefix tail
scoring, not the aligned full-response variant.

Overall, the additional latency of TAPD primarily reflects the cost of
obtaining tokenizer-independent localized feedback. This overhead is
typical of on-policy speech optimization, where generated trajectories
must be decoded and evaluated online. TAPD limits this cost by using
only one student rollout per example, unlike group-relative methods
whose rollout and teacher-evaluation costs scale with the number of
sampled responses.

\subsection{End-to-End Training Time}
\label{app:sec:end_to_end_training_time}

We additionally report the end-to-end wall-clock training time under
the shared main-experiment configuration. Each method is trained on
the same 10,000-example dataset using eight NVIDIA A800 80\,GB GPUs.
The reported times include rollout generation, teacher evaluation,
auxiliary SFT computation, backpropagation, and optimizer updates.

A representative Sentence-Level TTS-NLL run requires approximately
4 hours. Under the general cumulative-prefix implementation, TAPD
requires approximately 8 hours, while TTS-GRPO with \(G=3\) requires
approximately 10 hours. Thus, the end-to-end training cost of TAPD is
approximately \(2.0\times\) that of the sentence-level baseline, but
remains lower than the \(2.5\times\) cost of TTS-GRPO.

The end-to-end ratios are smaller than the isolated
feedback-construction ratios because all methods share substantial
computation from auxiliary SFT, backpropagation, and parameter updates.
Although cumulative scoring introduces additional latency, TAPD
requires only one student rollout per example. In contrast, the cost
of TTS-GRPO scales with the number of sampled trajectories, and even
the relatively small group size \(G=3\) used in our experiments results
in a longer overall training time.

\section{Robustness across Training Seeds}
\label{app:sec:training_seed_robustness}

To evaluate the robustness of TAPD to training randomness, we repeat
the CosyVoice3-based TAPD experiment using three training seeds:
\(42\), \(52\), and \(62\). Due to the cost of API-based evaluation,
it is impractical to conduct multi-seed evaluation for every teacher
configuration and baseline. We therefore focus this analysis on the
CosyVoice3-based TAPD setting, which serves as the primary TAPD
configuration reported in the main paper.

All runs start from the same Step-Audio-2-mini checkpoint and use the
same 10,000-example training set, CosyVoice3 teacher, optimization
hyperparameters, chunk configuration, and training duration. Only the
training random seed is changed. To isolate training-time variation
from decoding-time randomness, all resulting checkpoints are evaluated
using the same inference configuration and the fixed decoding seed
\(42\).

\begin{table*}[tp]
\centering
\scriptsize
\setlength{\tabcolsep}{3.2pt}
\renewcommand{\arraystretch}{1.10}
\caption{TAPD performance across three training seeds in the
CosyVoice3 setting. Avg.\ is the unweighted mean over the emotion,
speed, volume, and composite subsets. Seed \(42\) is the run reported
in the main paper.}
\label{app:tab:training_seed_robustness}
\begin{tabular}{
    l
    ccccc
    ccccc
}
\toprule
&
\multicolumn{5}{c}{\textbf{MOS \(\uparrow\)}} &
\multicolumn{5}{c}{\textbf{IFR (\%) \(\uparrow\)}} \\
\cmidrule(lr){2-6}
\cmidrule(lr){7-11}
\textbf{Training seed}
& \textbf{Emotion}
& \textbf{Speed}
& \textbf{Volume}
& \textbf{Composite}
& \textbf{Avg.}
& \textbf{Emotion}
& \textbf{Speed}
& \textbf{Volume}
& \textbf{Composite}
& \textbf{Avg.} \\
\midrule

\textbf{42 (main)}
& 3.748
& 4.532
& 4.455
& 4.205
& 4.235
& 43.29
& 86.67
& 73.67
& 72.00
& 68.90 \\

52
& 3.835
& 4.475
& 4.400
& 4.140
& 4.213
& 50.57
& 85.00
& 75.67
& 68.00
& 69.81 \\

62
& 3.686
& 4.490
& 4.443
& 3.985
& 4.151
& 45.71
& 84.33
& 70.33
& 61.00
& 65.35 \\

\midrule
Mean
& 3.756
& 4.499
& 4.433
& 4.110
& 4.200
& 46.52
& 85.33
& 73.22
& 67.00
& 68.02 \\

Std.
& 0.075
& 0.030
& 0.029
& 0.113
& 0.043
& 3.71
& 1.21
& 2.70
& 5.57
& 2.36 \\

\bottomrule
\end{tabular}
\end{table*}

Across the three independently trained checkpoints, TAPD achieves an
average MOS of \(4.200 \pm 0.043\) and an average IFR of
\(68.02\% \pm 2.36\) percentage points, where the uncertainty denotes
the sample standard deviation across training seeds. The average MOS
ranges from \(4.151\) to \(4.235\), while the average IFR ranges from
\(65.35\%\) to \(69.81\%\).

Importantly, even the lowest-performing seed achieves an average IFR
of \(65.35\%\), which remains higher than the strongest baseline in
the main comparison. Thus, the performance advantage of TAPD is
preserved across all three independently trained checkpoints rather
than being attributable to a single favorable initialization or
training trajectory.

The variation is not uniform across instruction categories.
Speaking-rate instruction following is particularly stable, with an
IFR standard deviation of only \(1.21\) percentage points. Greater
variation is observed for the composite subset, suggesting that
instructions involving multiple paralinguistic attributes are more
sensitive to training randomness than directly measurable attributes
such as speaking rate.

Seed \(52\) obtains the highest average IFR of \(69.81\%\), exceeding
the \(68.90\%\) result obtained by seed \(42\). Nevertheless, we retain
the seed-\(42\) checkpoint in the main paper because seed \(42\) is
used consistently for all compared methods. We do not replace the main
TAPD result with the higher-scoring seed-\(52\) checkpoint after
observing the evaluation outcomes. This fixed-seed protocol prevents
post-hoc seed selection and ensures that the comparison among SFT,
sentence-level optimization, GRPO, and TAPD is conducted under the
same random-seed setting.

Overall, the main-paper result lies close to the three-seed mean, is
not the best-performing run, and remains representative of the
observed performance range. Together with the fact that all three
TAPD runs outperform the strongest baseline, these results support the
robustness of TAPD's improvement to training-seed variation.

\section{Prompts for Style Conversion and Automatic Evaluation}
\label{app:sec:prompts}

This section provides the prompts used for offline style-instruction
construction and automatic evaluation. We include the complete prompts
to facilitate exact reproduction of our data-processing and evaluation
pipelines.

The MOS and IFR prompts are adopted directly from the official
UltraVoice evaluation setup~\cite{DBLP:journals/corr/abs-2510-22588} without modification.
Following the terminology of UltraVoice, we refer to its scalar
\(1\)--\(5\) instruction-alignment score as MOS. The style-conversion
prompt is used to extract a concise TTS instruction from each original
UltraVoice instruction.

\subsection{Style-Instruction Conversion Prompt}
\label{app:sec:style_conversion_prompt}

The original UltraVoice instruction may contain both response-content
requirements and acoustic speaking requirements. Because the
controllable TTS teacher should receive only the latter, we use
Qwen3-32B to convert each original instruction into a concise
style-only TTS instruction.

The conversion is performed offline once for every training example,
and the resulting instruction is cached. Importantly, the same cached
style instruction is used for both controllable-TTS synthesis of the
auxiliary SFT target and TTS-teacher conditioning during TAPD
on-policy scoring. We do not invoke a separate style parser or use a
different prompt for the two training branches.

\paragraph{System prompt.}

\begin{PromptBlock}
You are a TTS instruction converter.
Convert only the speaking requirements in instruction_text into one short TTS instruction.

Rules:
1. Focus only on speaking requirements, such as emotion, tone, volume, speaking rate, accent, speaking style, or vocal delivery. Do not include response content.
2. Output only the TTS instruction itself. Do not add explanations, labels, quotes, numbering, or extra details.
3. Only strengthen emotion, speaking rate, or volume when instruction_text explicitly specifies that dimension.
4. When strengthening a specified dimension, preserve its original category and direction: happy stays happy, sad stays sad, slow stays slow, fast stays fast, quiet stays quiet, and loud stays loud.
5. Do not add emotion, speed, or volume if they are not explicitly specified in instruction_text. Preserve other controls such as accent, style, or voice quality without intensifying them.
6. Use strong but natural modifiers when strengthening: very, deeply, strongly, extremely, intensely, very slowly, very quickly, very softly, or very loudly.
7. Keep the output concise, direct, and low-ambiguity.
8. If instruction_text is Chinese, output Chinese. If it is English or any other language, output English.
9. If no speaking requirement is given, output only a brief neutral instruction.

English examples:
Say this sentence in a very happy and joyful tone.
Say this sentence with deep sadness.
Say this sentence with strong anger.
Say this sentence very slowly and clearly.
Say this sentence at a very fast pace.
Say this sentence in a very quiet voice.
Say this sentence very loudly and clearly.
Say this sentence naturally.

Chinese examples:
用非常开心、充满喜悦的语气说这句话。
用深深悲伤的语气说这句话。
用强烈愤怒的语气说这句话。
非常缓慢、清晰地说这句话。
用非常快的语速说这句话。
用非常轻的声音说这句话。
用非常大的音量、清晰地说这句话。
自然地说这句话。
\end{PromptBlock}

\paragraph{User prompt template.}

\begin{PromptBlock}
instruction_text:
{instruction_text}

TTS instruction:
\end{PromptBlock}

The prompt explicitly removes response-content requirements to prevent
the TTS instruction from leaking semantic targets into the acoustic
conditioning signal. It strengthens emotion, speaking rate, or volume
only when the corresponding dimension is explicitly requested, while
preserving other controls such as accent, speaking style, and voice
quality. The output language follows the language of the original
instruction.

\subsection{MOS Evaluation Prompt}
\label{app:sec:mos_evaluation_prompt}

We use the official UltraVoice MOS prompt without modification. The
evaluator receives the generated waveform and the original user
instruction. It assesses the overall alignment of the audio with all
specified requirements, including spoken content and applicable
acoustic controls. The output is restricted to a single score between
\(1.0\) and \(5.0\), rounded to the nearest \(0.5\).

\begin{PromptBlock}
Your task is to evaluate the alignment between the provided audio (a model's speech reply as a .wav file) and the given instruction. The instruction typically includes the content to be spoken (e.g., a question) and various style controls (e.g., emotion, speed, volume, accent, style description, language change).

The evaluation should focus on how well the audio reply adheres to ALL aspects of the provided instruction. This includes, but is not limited to:
- Accuracy of the spoken content based on the "question" or core message in the instruction.
- Consistency with the specified emotion, if any.
- Adherence to the specified speed (e.g., fast, slow, normal), if any.
- Adherence to the specified volume (e.g., loud, soft, normal), if any.
- Correctness and consistency of the specified accent (e.g., British English, American English, specific regional accent), if any.
- Realization of the style description (e.g., "energetic and friendly," "formal and serious"), if any.
- Correct execution of any language change instruction (e.g., "switch to French for the last sentence"), if any.

Rate the audio on a scale of 1 to 5 based on the following criteria:
- 1 point: The audio completely fails to follow the instruction. The spoken content may be incorrect, significantly incomplete, or unintelligible. Multiple specified style controls are ignored, misapplied, or contradictory to the instruction.
- 2 points: The audio attempts to follow parts of the instruction but does so poorly or inconsistently. There may be major deviations in spoken content, or several style controls are noticeably incorrect, faint, mismatched, or missing. The overall result significantly deviates from the instruction.
- 3 points: The audio generally follows the main aspects of the instruction (e.g., content is mostly accurate, dominant style controls are attempted). However, there are noticeable inconsistencies in one or more style controls, some secondary style controls are not adequately met, or the overall delivery has clear flaws in matching the full instruction.
- 4 points: The audio effectively follows most aspects of the instruction with only minor imperfections or slight deviations in one or two style control elements. The spoken content is accurate, and the overall delivery strongly aligns with the instruction. Most specified style controls are well-realized.
- 5 points: The audio perfectly matches all aspects of the provided instruction. The spoken content is accurate. All specified style controls (emotion, speed, volume, accent, style description, language change, etc., as applicable) are flawlessly, clearly, and appropriately executed, resulting in a natural and fully compliant delivery.

Please note: Your evaluation should be independent and strictly based on the provided instruction and the audio's alignment with it. Consider all specified parameters in the instruction.

Please provide your score based on the audio's adherence to ALL specified elements in the instruction.

Below is the transcription of user's instruction:
[Instruction Details]
{instruction}

After evaluating, please output ONLY the final calculated score (a number between 1.0 and 5.0, rounded to the nearest 0.5) without anything else.

Please strictly follow the standards and avoid leniency in your evaluation. Ensure that the score reflects the exact alignment between the audio and the full instruction, without overestimating or underestimating the quality.
\end{PromptBlock}

\subsection{IFR Evaluation Prompt}
\label{app:sec:ifr_evaluation_prompt}

We also use the official UltraVoice IFR prompt without modification.
Unlike the MOS prompt, IFR evaluates only whether all requested
acoustic controls are followed. It explicitly ignores response-content
accuracy and produces a binary output. An output of \(1\) requires
every specified control to be correctly realized; failure to satisfy
any requested control results in an output of \(0\).

\begin{PromptBlock}
Your task is to determine whether the provided audio strictly follows the acoustic control instructions in the given prompt. Focus ONLY on the acoustic aspects and output ONLY 1 (follows instruction) or 0 (does not follow instruction).

The evaluation should check for:
1. Emotion Control (if specified): Does the voice express the exact requested emotion?
2. Speed Control (if specified): Does the speech maintain the exact requested pace?
3. Volume Control (if specified): Does the audio maintain the exact requested volume level?
4. Accent Control (if specified): Does the voice use the exact requested accent?
5. Style Description (if specified): Does the voice match the exact style descriptors?
6. Language Switch (if specified): Does the voice switch to the requested language at the specified point?

IMPORTANT:
- Output ONLY 1 or 0
- 1 = ALL specified controls are correctly followed
- 0 = ANY specified control is not followed correctly
- Ignore content accuracy completely
- Consider ONLY the specified control elements

Below is the instruction with acoustic controls:
[Instruction Details]
{instruction}

After evaluating, output ONLY 1 or 0.
\end{PromptBlock}

For both metrics, the evaluator output is parsed only when it matches
the required numeric format. No additional natural-language reasoning
or evaluator-generated explanation is incorporated into the reported
score. All compared systems are evaluated using the same prompt,
evaluator configuration, and parsing procedure.

\section{Publication Status and Code Availability}
\label{app:sec:code_availability}

This manuscript has not yet been accepted for publication.
The accompanying code may be subject to further modifications.
We will make the code publicly available upon acceptance of the paper.

\end{CJK*}
\end{document}